\documentclass[11pt]{article}
\usepackage{listings}

\usepackage[final]{acl}
\usepackage{booktabs}
\usepackage{multirow}
\usepackage{amssymb}
\usepackage{fancyvrb}
\usepackage[most]{tcolorbox}
\usepackage{times}
\usepackage{latexsym}
\usepackage{svg}
\usepackage[T1]{fontenc}

\usepackage[utf8]{inputenc}

\usepackage{microtype}

\usepackage{inconsolata}

\usepackage{graphicx}

\usepackage{amsmath,amsfonts}
\usepackage{algorithmic}
\usepackage{array}
\usepackage{textcomp}
\usepackage{stfloats}
\usepackage{url}
\usepackage{verbatim}
\usepackage{booktabs}
\usepackage{tabularx}
\usepackage{multirow}
\usepackage{hyperref}
\usepackage{arydshln}
\usepackage{comment}
\usepackage{booktabs}
\usepackage[table,dvipsnames]{xcolor}
\usepackage{lipsum}
\usepackage{subcaption}
\usepackage{float}
\usepackage{changepage}
\usepackage{enumitem}    
\usepackage{amssymb}     
\usepackage{pifont}
\title{SEEK: Semantic Evidence Extraction via Adaptive ChunKing for Multilingual Fact-Checking}

\author{
Gaurav Kumar *, Babu Kumar *,Ayush Garg, Aditya Kishore, Jasabanta Patro \\
   Department of Data Science and Engineering \\
  Indian Institute of Science Education and Research, Bhopal, India \\
  \texttt{\{babu21, gaurav22, ayushg24, adityak21, jpatro\}@iiserb.ac.in}\\
}

\begin{document}
\maketitle

\begin{abstract}

Multilingual fact verification requires evidence that is both relevant and sufficiently complete for reliable factuality prediction. However, existing systems often rely on search snippets, sentence-level evidence, or locally segmented passages, which can miss decisive context and produce fragmented evidence. To overcome these limitations, we propose \textit{\textsc{SEEK}}, a \textit{S}emantic \textit{E}vidence \textit{E}xtraction with Adaptive chun\textit{K}ing framework that constructs coherent evidence chunks from full fact-checking articles by identifying semantic topic transitions and preserving local verification context. The constructed chunks are encoded with a multilingual encoder, and multilingual LLMs are then fine-tuned with LoRA adapters for veracity prediction. Experiments on X-FACT and \textsc{RU22Fact} show that \textsc{SEEK} improves Macro-F1 by up to \textbf{10}\% over semantic chunking,\textbf{ 19}\% over sentence chunking, and \textbf{20}\% over search-snippet baselines. Evidence completeness and significance analyses further show that \textsc{SEEK} preserves richer verification context and enables more reliable multilingual fact-checking.
\end{abstract}

\section{Introduction}

\noindent The multilingual nature of today’s online information ecosystem has made misinformation increasingly difficult to detect, contextualize, and verify~\cite{panchendrarajan2024claim}. A single misleading claim may circulate across languages, platforms, and regional communities, often appearing in modified forms through translation, paraphrasing, or culturally specific framing~\cite{quelle2025lost,peng-etal-2025-semeval}. In such settings, fact verification is not limited to deciding whether a claim is true or false; it also requires identifying reliable evidence, preserving the meaning of the claim across languages, and reasoning over heterogeneous sources that may differ in structure, language, and level of detail~\cite{guo-etal-2022-survey,gupta-srikumar-2021-x}.

\begin{figure}[t]
    \centering
    \includegraphics[width=\columnwidth]{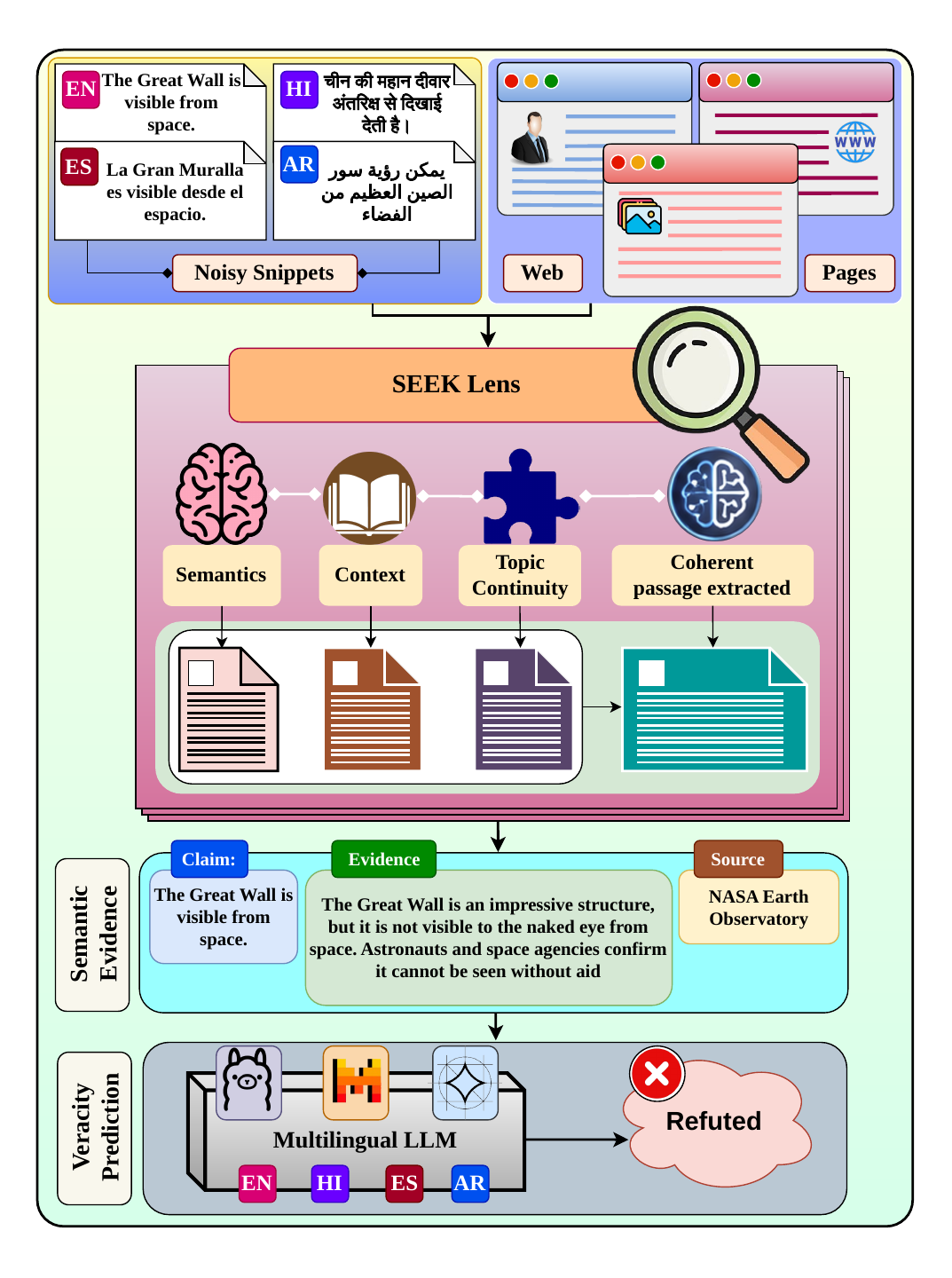}
    \caption{High-level illustration of the proposed multilingual fact verification framework. Noisy multilingual snippets and full web pages are transformed through a SEEK lens into coherent evidence, which is then used by a multilingual LLM for veracity prediction.}
    \label{fig:intro_gist}
     \vspace{-15pt}
\end{figure}

\noindent Human fact-checkers remain central to this process because they can interpret context, compare sources, and evaluate evidence with linguistic and domain-specific judgment~\cite{ijcai2021p619}. However, manual fact-checking is difficult to scale against the volume and speed of online misinformation~\cite{ijcai2021p619,nanhekhan2025flashcheck}. This has motivated automated fact verification, where NLP systems retrieve relevant evidence and predict claim veracity~\cite{guo-etal-2022-survey,zheng-etal-2024-evidence}. Despite recent progress, system reliability depends strongly on evidence quality~\cite{zheng-etal-2024-evidence,nanhekhan2025flashcheck}, especially in multilingual settings where relevant information may be scattered across long fact-checking articles, news reports, official documents, or web pages~\cite{zheng-etal-2024-evidence,gupta-srikumar-2021-x,cekinel-etal-2024-cross,peng-etal-2025-semeval,NEURIPS2023_cd86a305}. Such sources often mix background discussion, quoted claims, contextual explanations, and final verdicts within the same document~\cite{augenstein-etal-2019-multifc,NEURIPS2023_cd86a305}. As a result, short snippets or isolated sentences may miss the complete verification context, whereas full-document processing can introduce substantial irrelevant information~\cite{zheng-etal-2024-evidence,NEURIPS2023_cd86a305}. This creates a mismatch between the retrieved evidence unit and the evidence span actually required for reliable claim verification~\cite{zheng-etal-2024-evidence,NEURIPS2023_cd86a305}.

\noindent Existing evidence retrieval strategies often rely on fixed-size passages, sentence-level segmentation, or externally retrieved snippets~\cite{chen2022gere,NEURIPS2023_cd86a305}. While simple and efficient, these approaches may not align with the semantic structure of fact-checking documents~\cite{zhang-etal-2023-relevance}. Fixed windows can split reasoning across boundaries, whereas sentence-level retrieval may discard the surrounding context needed for correct interpretation~\cite{zhang-etal-2023-relevance,zheng-etal-2024-evidence}. This limitation is amplified in multilingual settings, where translation variation, code-mixing, and language-specific discourse patterns can weaken claim-evidence alignment~\cite{panchendrarajan2024claim,peng-etal-2025-semeval}.

\noindent To address these limitations, we propose \textsc{SEEK}, a Semantic Evidence Extraction with adaptive chunKing framework for multilingual fact verification. Rather than relying on short snippets or fixed-length passages, \textsc{SEEK} constructs coherent evidence chunks from full web documents by detecting semantic topic shifts in a shared multilingual embedding space. These chunks preserve complete verification context while reducing irrelevant noise, and are then retrieved for veracity prediction using LoRA-tuned~\cite{hu2022lora} multilingual large language models. The central idea is that reliable multilingual fact verification requires not only strong language models, but also better evidence granularity. Improving how evidence is segmented and retrieved can therefore enhance the reliability of downstream veracity prediction across multilingual settings.

\noindent \textbf{Our contributions are as follows:}
\begin{itemize}
  
    \item We introduce \textsc{SEEK}, a multilingual evidence construction framework that combines contextual topic-shift detection, score smoothing, adaptive thresholding, and boundary overlap to generate verification-oriented evidence chunks.
    
    \item We combine \textsc{SEEK} with multilingual dense retrieval and LoRA-tuned multilingual LLMs, achieving strong veracity prediction performance across multilingual fact-checking benchmarks.

    \item We perform comprehensive evidence analysis on \textsc{X-FACT} and \textsc{RU22Fact}, including evidence completeness, similarity, and statistical significance studies across multilingual and generalization settings.

    \item We further conduct translation-based evaluation to analyze the effect of language normalization on evidence retrieval and veracity prediction in multilingual fact verification.
\end{itemize}

\section{Related Work}

\subsection{General and Multilingual Fact-Checking}

Automated fact-checking is typically framed as verifying a claim using either model-internal knowledge or external evidence. Benchmarks such as FEVER~\cite{thorne2018fever} and LIAR~\cite{wang2017liar} established evidence-based and fine-grained fact verification in English, while multilingual datasets such as X-FACT~\cite{gupta2021xfact} and \textsc{RU22Fact}\cite{zeng2024ru22fact} extend this task across languages, domains, and resource settings. These benchmarks show that multilingual fact verification depends not only on strong veracity models, but also on evidence that is relevant, complete, and linguistically usable for the target claim.

\subsection{Evidence Extraction and Retrieval}

\textbf{Snippet- and sentence-level retrieval.}
Many fact-checking systems rely on search snippets, individual sentences, or short passages as evidence~\cite{gupta-srikumar-2021-x}. 
These units are efficient and focused, but they often fragment the verification context. 
In multilingual settings, this issue is amplified because snippets may omit key background, distort named entities, or miss the link between the claim, investigation, and verdict~\cite{gupta-srikumar-2021-x}. 
Dense retrievers such as DPR and multilingual-e5 improve semantic matching between claims and evidence candidates~\cite{karpukhin-etal-2020-dense,wang2024multilingual}. Similarly, CONCRETE improves cross-lingual fact-checking by learning multilingual retrieval representations from trusted multilingual corpora~\cite{huang-etal-2022-concrete}. However, these retrieval methods do not directly resolve the granularity problem: the retrieved unit may still be too short to contain complete verification context or too broad to avoid irrelevant information.

\noindent \textbf{Document chunking methodologies.}
Other approaches retrieve evidence from longer documents by splitting them into fixed-size, sentence-aware, or semantic chunks~\cite{qu2025semantic,kiss2025maxmin}. 
Fixed and sentence-based chunking are simple, but they can cut off decisive information when verification cues appear beyond the boundary~\cite{qu2025semantic}. 
Semantic chunking improves coherence, yet local similarity-based boundaries can still be unstable in noisy multilingual fact-checking articles that mix claims, explanations, quotes, and verdicts~\cite{qu2025semantic,kiss2025maxmin}. 
This creates a key trade-off: shorter units reduce noise but risk context fragmentation, while longer units preserve context but may introduce irrelevant information~\cite{qu2025semantic}.

\subsection{Evidence Granularity for Multilingual Verification}

Prior work has advanced multilingual fact-checking through better datasets, cross-lingual retrieval methods such as CONCRETE, translation-based normalization, and LLM-based verification~\cite{huang-etal-2022-concrete,peng2025semeval}. 
However, the role of evidence granularity remains underexplored: retrieved text must not only be semantically related to the claim, but also complete enough to support factuality prediction~\cite{viswanathan2025claimiq}. 
This is especially important for multilingual verification, where systems must handle language variation, low-resource settings, noisy web pages, and dispersed evidence cues~\cite{peng2025semeval,viswanathan2025claimiq}.
This gap motivates an evidence construction strategy that operates beyond isolated snippets or fixed windows while still avoiding overly broad document segments. By aligning chunk boundaries with contextual topic shifts and retaining boundary overlap, the proposed approach preserves the continuity between the claim, supporting details, and verdict. In this way, it reduces context fragmentation in evidence retrieval and provides more complete verification evidence for multilingual fact-checking.

\section{Datasets Details}
In this section, we report the details of datasets used to evaluate our study. Since our work focuses on multilingual fact-checking, we consider two multilingual benchmarks: \textbf{X-FACT}~\cite{gupta2021xfact} and \textbf{\textsc{RU22Fact}}~\cite{zeng2024ru22fact}. These datasets allow us to evaluate the proposed framework across diverse languages, claims, and evidence sources. The statistical details of both datasets are summarized in Table~\ref{tab:dataset_stats}. Additional dataset descriptions are provided in Appendix ~\ref{App:DatasetDetails}.

\begin{table}[t]
\centering
\small
\setlength{\tabcolsep}{3pt} 
\renewcommand{\arraystretch}{1.15}

\resizebox{\columnwidth}{!}{%
\begin{tabular}{l@{\hspace{20pt}}c c r r r r}
\toprule
\textbf{Datasets} & \textbf{No. of Classes} & \textbf{Language} &
\textbf{Train} & \textbf{Val} & \textbf{Test} & \textbf{Total} \\
\midrule

\multicolumn{7}{l}{\textsc{Multilingual}} \\
~~~-X-FACT    & 7 & 25 languages  & 19,079 & 2,535 & 9,575 & 31,189 \\
~~~-\textsc{RU22Fact}  & 3 & 4 languages   & 11,217 & 1,600 & 3,216 & 16,033 \\

\bottomrule
\end{tabular}%
}

\caption{Details of multilingual datasets considered in our work. }
\label{tab:dataset_stats}
\end{table}

\section{Methodology}
\label{methodology}
The overall fact-checking pipeline used in this work consists of four key components: (i) a web crawler, (ii) the \textsc{SEEK} chunking module, (iii) a multilingual dense retriever, and (iv) instruction-tuned large language models, namely LLaMA, Gemma, and Mistral. The individual components are described below:

\noindent \textbf{Web crawler:} In the X-FACT dataset~\cite{gupta2021xfact}, each claim is accompanied by five Google Search snippets and their corresponding source URLs. Since these snippets are often insufficient for reliable verification, we crawl the full content of each URL using \textbf{Crawl4AI}~\cite{crawl4ai2024}. The crawler removes boilerplate elements such as navigation bars and scripts and extracts the main textual content, yielding up to five documents per claim, with lengths ranging from hundreds to several thousand tokens. In contrast, each claim in \textsc{RU22Fact} is linked to a single source URL, which we retrieve and clean using the same pipeline.

\noindent \textbf{Chunking Module:}
Due to the long and unstructured nature of crawled web pages, the extracted documents are divided into smaller textual units before evidence retrieval. Effective chunking should balance efficiency and completeness: very short chunks may miss the context needed for verification, while very long chunks may introduce irrelevant content. Existing sentence-based or semantic chunking methods can still produce incomplete evidence when verification cues are spread across nearby but separated parts of a document. To address this, \textsc{SEEK} uses context-window topic-shift detection, smoothing, adaptive thresholding, and overlap-aware chunk construction to form coherent evidence passages.As shown in Appendix Figure~\ref{fig:chunking_example}, this helps preserve both the viral claim context and the later refuting evidence within the same retrieval unit.

\paragraph{Baseline chunking methods.}
We compare \textsc{SEEK} with two common document chunking strategies. Sentence-aware fixed-size chunking divides a document into consecutive groups of complete sentences under a fixed token budget of 512 tokens. This method preserves sentence boundaries and avoids cutting sentences in the middle, but it does not consider topic changes within the document. Therefore, it may either stop before the full verification context is captured or combine unrelated sentences in the same chunk. Semantic chunking \cite{langchain_semantic_chunking}, in contrast, uses a multilingual sentence encoder to map each sentence $s_i$ into an embedding $\mathbf{e}_i=f(s_i)$ and places a boundary when the cosine similarity between neighboring sentence embeddings falls below a threshold $\tau$. Although this captures local semantic changes, it mainly relies on adjacent sentence similarity and can still produce incomplete or noisy evidence when the verification context spans multiple sentences.

\paragraph{SEEK.}Although standard semantic chunking is more meaningful than fixed-size splitting, it can still be unstable for noisy multilingual web documents. A boundary decision based only on adjacent sentences is sensitive to local variations, translation style, and short sentence noise. To address this limitation, we propose a strategy, SEEK, that detects topic shifts using context-window comparison rather than only sentence-to-sentence similarity.

\noindent For each possible boundary position $i$, we define a left context window and a right context window:
\[
L_i = \{s_{i-w+1}, \ldots, s_i\},
\quad
R_i = \{s_{i+1}, \ldots, s_{i+w}\},
\]
where $w$ is the window size (set to $w=3$ in our experiments). The embeddings of these two windows are computed by averaging sentence embeddings:
\[
\mathbf{l}_i = \frac{1}{|L_i|}\sum_{s_j \in L_i} \mathbf{e}_j,
\quad
\mathbf{r}_i = \frac{1}{|R_i|}\sum_{s_j \in R_i} \mathbf{e}_j.
\]

\noindent We then compute a semantic shift score:
\[
\Delta_i = 1 - \cos(\mathbf{l}_i, \mathbf{r}_i).
\]

\noindent A higher value of $\Delta_i$ indicates a stronger semantic change between the left and right contexts, suggesting a possible topic boundary. To reduce noisy fluctuations, the shift scores are smoothed:
\[
\tilde{\Delta}_i = \frac{1}{k}\sum_{j=i-\lfloor k/2 \rfloor}^{i+\lfloor k/2 \rfloor} \Delta_j,
\]
where $k$ (set to $k=3$ in our experiments) is the smoothing window size.

\noindent Instead of using a fixed threshold, we apply adaptive thresholding based on the distribution of shift scores within the document:
\[
\tau = \text{Percentile}(\tilde{\Delta}, p),
\]
where $p$ controls the selectivity of boundary detection and is set to $p=95\%$ in our experiments. A boundary is selected when:
\[
b_i =
\begin{cases}
1, & \text{if } \tilde{\Delta}_i \geq \tau, \\
0, & \text{otherwise}.
\end{cases}
\]
\begin{figure*}[ht!]
    \centering
    \makebox[\textwidth][c]{%
        \includegraphics[width=\textwidth]{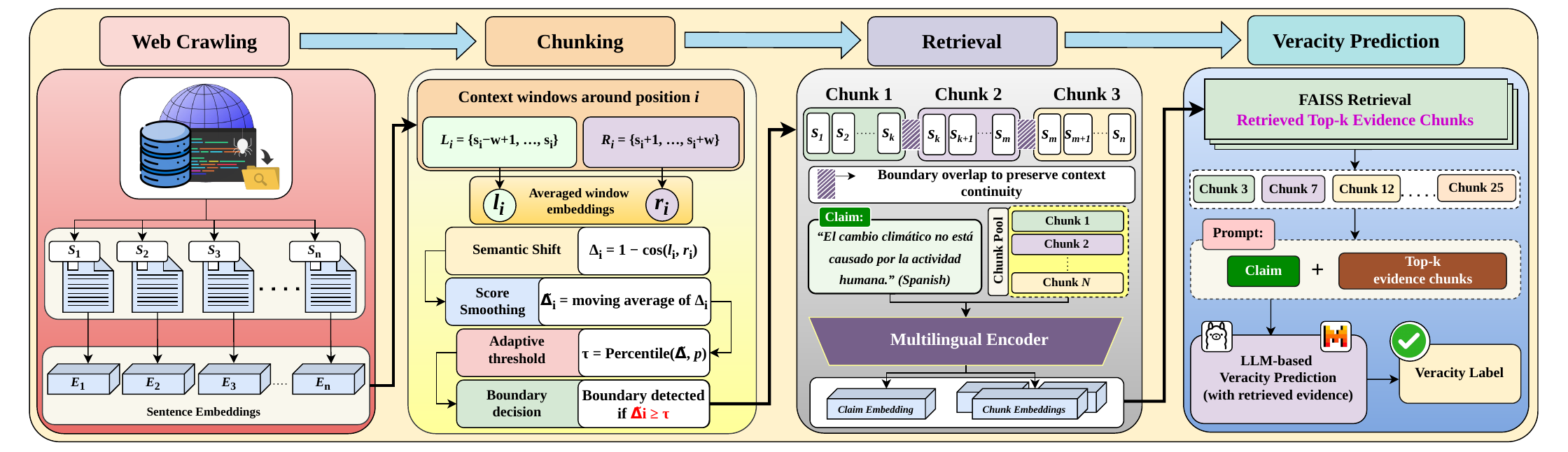}
    }
    \caption{Overview of the SEEK and retrieval-based veracity prediction pipeline.}
    \label{fig:improved_semantic_chunking}
\end{figure*}
\noindent Finally, sentences between selected boundaries are grouped into chunks under a maximum token budget. We also include a small overlap between consecutive chunks to preserve boundary-level context. This produces chunks that are semantically coherent, token-efficient, and more suitable for dense evidence retrieval in multilingual fact-checking. An example visualization of the raw semantic shift scores, smoothing process, and adaptive boundary selection is provided in \textbf{Appendix~\ref{app:chunking_example}}.

\noindent \textbf{Multilingual dense retriever:}
After generating semantically coherent chunks with \textsc{SEEK}, we retrieve claim-relevant evidence using a multilingual dense retrieval pipeline. Each claim and candidate chunk are encoded with \textbf{multilingual-e5-large-instruct}~\cite{wang2024multilingual}, which maps multilingual claims and evidence passages into a shared semantic embedding space.

\noindent Given a claim $c$ and crawled source documents $\mathcal{D}=\{d_1,d_2,\ldots,d_n\}$, each document $d_j$ is segmented into a set of chunks:
\[
\mathcal{P}_j = \{p_{j1}, p_{j2}, \ldots, p_{jm}\}.
\]
The claim and chunk embeddings are computed as:
\[
\mathbf{h}_c = \text{Encoder}(c),
\qquad
\mathbf{h}_{p_{ji}} = \text{Encoder}(p_{ji}),
\]
where both are produced using the same multilingual encoder. Following the instruction-tuned formulation of multilingual-e5, the claim is encoded with the retrieval instruction:
\begin{quote}
\textit{Instruct: Given a claim, retrieve relevant evidence from web documents that support or refute the claim.}
\end{quote}

\noindent All chunks from the crawled documents are pooled into a global multilingual evidence space and indexed with FAISS~\cite{douze2024faiss}. For each claim, cosine similarity is computed between the claim embedding and every candidate chunk:
\[
s_{ji} = \cos(\mathbf{h}_c,\mathbf{h}_{p_{ji}}).
\]
Instead of retrieving evidence separately from each document, we perform global retrieval across the complete chunk pool. We first retrieve the top-$N$ candidates using FAISS, with $N=20$:
\[
\mathcal{R} = \operatorname{TopN}\left(
\cos(\mathbf{h}_c,\mathbf{h}_{p_{ji}})
\right).
\]
These candidates are then re-ranked using the same multilingual bi-encoder, and the final top-$K$ chunks are selected, with $K=5$:
\[
\mathcal{E}^* = \operatorname{TopK}(\mathcal{R}).
\]
Here, $\mathcal{E}^*$ denotes the final evidence set passed to the downstream veracity prediction model. The selected evidence chunks are concatenated with the claim for veracity prediction. Unlike sentence-level retrieval, which may return isolated or incomplete evidence, \textsc{SEEK} retrieves semantically coherent chunks that preserve topic continuity and verification context while reducing irrelevant noise.

\noindent \textbf{Instruction-tuning of large language models:}
After retrieval, the top-$K$ ranked evidence chunks $\mathcal{E}^{*}=\{p_1^*,p_2^*,\ldots,p_K^*\}$ are used for downstream veracity prediction. For each claim, the input is formed by concatenating the task instruction, claim, and retrieved evidence:
\[
\text{Input} = \mathcal{I} \oplus \mathcal{C} \oplus p_1^* \oplus p_2^* \oplus \cdots \oplus p_K^* ,
\]
where $\mathcal{I}$ denotes the dataset-specific instruction, $\mathcal{C}$ is the claim, and $\oplus$ denotes textual concatenation. The instructions for \textsc{X-FACT} and \textsc{RU22Fact} are provided in Appendix~\ref{app:instructions}. For \textsc{X-FACT}, evidence is retrieved from up to five associated web sources, while for \textsc{RU22Fact}, it is retrieved from the corresponding source article.

\noindent We formulate veracity prediction as an instruction-following causal language modeling task~\cite{zhang2025instructiontuninglargelanguage}, where the model generates the factuality label conditioned on the claim and evidence. We fine-tune multilingual LLMs, including \textbf{LLaMA}, \textbf{Gemma}, and \textbf{Mistral}, using \textbf{LoRA}~\cite{hu2022lora}. The base model is frozen, and only the low-rank adaptation parameters and output language modeling head are updated. Given the target label sequence $y_{1:T}$, the training loss is:
\[
\mathcal{L}
=
-\sum_{t=1}^{T}
\log p(y_t \mid y_{<t}, \text{Input}).
\]
This trains the model to generate the correct veracity label using the claim and the retrieved evidence, where \textsc{SEEK} provides complete verification context for prediction.
\section{Experimental Setup}

This section describes the baselines, implementation settings, and evaluation protocols used to assess both veracity prediction performance and retrieval quality in multilingual fact-checking.

\subsection{Baselines}

We compare \textsc{SEEK} with several retrieval and evidence construction baselines.

\noindent\textbf{Google Search Snippets:}
Following the original \textsc{X-FACT} setup, this baseline uses the provided Google Search snippets as evidence.

\noindent\textbf{CONCRETE Retriever:}
We compare against CONCRETE~\cite{huang-etal-2022-concrete}, a multilingual dense retriever trained with the Cross-lingual Inverse Cloze Task (X-ICT). It retrieves evidence passages using multilingual dense similarity matching.

\noindent\textbf{Sentence Chunking:}
This baseline splits each crawled webpage into consecutive sentence groups under a fixed token budget. This strategy is described in detail in Section~\ref{methodology}.

\noindent\textbf{Semantic Chunking:}
This baseline groups neighboring sentences into chunks based on embedding similarity. This strategy is also described in detail in Section~\ref{methodology}.

\noindent\textbf{LLM-Generated Evidence:}
For \textsc{RU22Fact}, we also compare with the LLM-generated evidence setting from the original \textsc{RU22Fact} work~\cite{zeng2024ru22fact}, where evidence is generated by an LLM instead of retrieved from crawled web pages.

\subsection{Implementation Details}

\noindent We perform multilingual dense retrieval using \textbf{intfloat/multilingual-e5-large-instruct}, which encodes claims and evidence chunks into a shared multilingual semantic space. The resulting chunk embeddings are indexed with FAISS to enable efficient dense similarity search. After retrieval, the top-ranked evidence chunks are concatenated with the input claim and provided to instruction-tuned large language models for veracity prediction. We fine-tune \textbf{LLaMA}, \textbf{Gemma}, and \textbf{Mistral} using Low-Rank Adaptation (LoRA) under a causal language modeling objective. Further details on these models are provided in Appendix~\ref{app:model_details}. All experiments are implemented using \textbf{LLaMAFactory}~\cite{zheng2024llamafactory} and conducted on a single NVIDIA A100 80 GB PCIe GPU.

\subsection{Translation-based Retrieval Analysis}

\noindent To study the effect of multilingual variation on retrieval, we translate all \textsc{X-FACT} and \textsc{RU22Fact} claims and crawled webpages into English before applying the same \textsc{SEEK} pipeline. The translated documents are chunked into semantically coherent passages, and retrieval is performed using \textbf{sentence-transformers/all-MiniLM-L6-v2}~\cite{allMiniLML6v2}. The retrieved chunks are then combined with the translated claim for veracity prediction. This setup tests whether retrieval improves when claim-evidence matching is performed in a unified English semantic space rather than a multilingual one.

\subsection{Evaluation Protocol}

For \textsc{X-FACT}, we report Macro-F1 under the original benchmark settings: in-domain (ID), out-of-domain (OOD), and zero-shot (ZS), with split details provided in Appendix~\ref{App:DatasetDetails}. For \textsc{RU22Fact}, we report Macro-F1 on the official test split. In addition to veracity prediction, we evaluate retrieval quality using claim-evidence semantic similarity, evidence completeness analysis, and qualitative retrieval analysis. To ensure fair comparison, we use the same fine-tuning, retrieval, decoding, and evaluation settings across all chunking strategies; detailed hyperparameters are provided in Appendix~\ref{app:hyperparameters}.
\section{Results and Discussion}
\newcommand{\std}[1]{\textnormal{\tiny\,($\pm$#1)}}
\begin{table}[t]
\centering
\footnotesize
\renewcommand{\arraystretch}{1.15}
\setlength{\tabcolsep}{3pt}

\begin{tabular}{l l c c c}
\toprule
\textbf{Setting} & \textbf{Model} & \textbf{ID} & \textbf{OOD} & \textbf{ZS} \\
\midrule

\multirow{3}{*}{\shortstack[l]{Search\\Snippets}}
 & LLaMA   & 0.40\std{1.07} & 0.26\std{0.25} & 0.22\std{1.04} \\
 & Gemma   & 0.43\std{1.63} & 0.28\std{1.00} & 0.20\std{1.36} \\
 & Mistral & 0.46\std{1.05} & 0.30\std{0.45} & 0.21\std{0.70} \\
\midrule

\multirow{3}{*}{CONCRETE}
 & LLaMA   & 0.33\std{3.90} & 0.19\std{0.18} & 0.18\std{1.19} \\
 & Gemma   & 0.35\std{1.60} & 0.21\std{1.42} & 0.15\std{0.54} \\
 & Mistral & 0.41\std{0.71} & 0.22\std{0.46} & 0.18\std{0.37} \\
\midrule

\multirow{3}{*}{\shortstack[l]{Sentence \\Chunking}}
 & LLaMA   & 0.65\std{0.40} & 0.31\std{1.31} & 0.22\std{0.91} \\
 & Gemma   & 0.61\std{0.30} & 0.35\std{1.63} & 0.23\std{1.16} \\
 & Mistral & 0.63\std{0.83} & 0.37\std{0.95} & 0.25\std{1.48} \\
\midrule

\multirow{3}{*}{\shortstack[l]{Semantic \\Chunking}}
 & LLaMA   & 0.63\std{0.99} & 0.33\std{0.65} & 0.23\std{1.40} \\
 & Gemma   & 0.61\std{0.63} & 0.37\std{0.44} & 0.23\std{1.17} \\
 & Mistral & 0.60\std{0.28} & 0.37\std{0.37} & 0.21\std{0.16} \\
\midrule

\multirow{3}{*}{\shortstack[l]{\textbf{SEEK}}}
 & LLaMA   & \textbf{0.67}\std{0.51} & \textbf{0.41}\std{1.48} & \textbf{0.30}\std{0.62} \\
 & Gemma   & \textbf{0.64}\std{2.54} & \textbf{0.39}\std{0.28} & \textbf{0.24}\std{0.40} \\
 & Mistral & \textbf{0.66}\std{0.55} & \textbf{0.39}\std{0.57} & \textbf{0.27}\std{1.79} \\
\bottomrule
\end{tabular}

\caption{Macro-F1 (mean $\pm$ std) on X-Fact across In-Domain (ID), Out-of-Domain (OOD), and Zero-Shot (ZS) settings.}
\label{tab:xfact_results}
\end{table}
\begin{table}[t]
\centering
\footnotesize
\renewcommand{\arraystretch}{1.12}
\setlength{\tabcolsep}{4pt}

\begin{tabularx}{\columnwidth}
{>{\raggedright\arraybackslash}p{0.32\columnwidth}
 >{\raggedright\arraybackslash}p{0.28\columnwidth}
 >{\centering\arraybackslash}X}
\toprule
\textbf{Setting} & \textbf{Model} & \textbf{Macro-F1} \\
\midrule

\multirow{3}{*}{LLM}
 & LLaMA   & 0.73\std{0.47} \\
 & Gemma   & 0.70\std{2.53} \\
 & Mistral & 0.72\std{0.71} \\
\midrule

\multirow{3}{*}{\shortstack[l]{Sentence\\Chunking}}
 & LLaMA   & 0.71\std{0.92} \\
 & Gemma   & 0.71\std{1.85} \\
 & Mistral & 0.72\std{1.02} \\
\midrule

\multirow{3}{*}{\shortstack[l]{Semantic\\Chunking}}
 & LLaMA   & 0.79\std{0.54} \\
 & Gemma   & 0.78\std{0.59} \\
 & Mistral & 0.79\std{0.67} \\
\midrule

\multirow{3}{*}{\shortstack[l]{\textbf{SEEK}}}
 & LLaMA   & \textbf{0.89}\std{0.37} \\
 & Gemma   & \textbf{0.90}\std{0.42} \\
 & Mistral & \textbf{0.90 }\std{0.37} \\
\bottomrule
\end{tabularx}

\caption{Macro-F1 on \textsc{RU22Fact} across four settings: LLM, sentence chunking,    semantic chunking, and SEEK.}
\label{tab:ru22fact_results}
\end{table}

We report veracity prediction results using Macro-F1 as the main evaluation metric. The results are summarized in Tables~\ref{tab:xfact_results} and~\ref{tab:ru22fact_results}.

\subsection{Findings on X-FACT}

\noindent Table~\ref{tab:xfact_results} reports Macro-F1 results on \textsc{X-FACT} across in-domain (ID), out-of-domain (OOD), and zero-shot (ZS) settings. \textsc{SEEK} consistently outperforms search-snippet evidence and the \textsc{CONCRETE} retriever, showing that veracity prediction benefits from more complete and semantically coherent evidence. The strongest results are obtained by \textbf{LLaMA with \textsc{SEEK}}, achieving \textbf{0.67} Macro-F1 in ID, \textbf{0.41} in OOD, and \textbf{0.30} in ZS. This corresponds to gains of up to \textbf{+0.21} over the best search-snippet baselines and up to \textbf{+0.26} over \textsc{CONCRETE}. These gains are especially important in OOD and ZS settings, where domain and language shifts make evidence retrieval more challenging. Across LLaMA, Gemma, and Mistral, \textsc{SEEK} yields stable improvements over sentence-level and standard semantic chunking baselines. This indicates that the gains arise mainly from improved evidence construction rather than a particular LLM backbone. Overall, coherent document-level chunks provide more reliable multilingual verification signals than snippets or isolated retrieval units.





\begin{table}[t]
\centering
\footnotesize
\renewcommand{\arraystretch}{1.12}
\setlength{\tabcolsep}{4pt}

\begin{tabularx}{\columnwidth}
{>{\raggedright\arraybackslash}p{0.32\columnwidth}
 >{\centering\arraybackslash}X
 >{\centering\arraybackslash}X
 >{\centering\arraybackslash}X}
\toprule
\textbf{Models} & \textbf{ID} & \textbf{OOD} & \textbf{ZS} \\
\midrule

LLaMA   & 0.65\std{2.35} & \textbf{0.41\std{0.69}} & 0.36\std{3.54} \\
Gemma   & 0.65\std{2.12} & 0.39\std{0.66} & \textbf{0.37\std{1.97}} \\
Mistral & \textbf{0.66\std{1.06}} & 0.40\std{0.63} & 0.37\std{1.02} \\

\bottomrule
\end{tabularx}

\caption{Translation-based Macro-F1 performance of \textsc{SEEK} on \textsc{X-Fact} using different multilingual LLMs. Results are reported as Macro-F1 (mean $\pm$ std) across random seeds.}
\label{tab:cross_domain_results}
\end{table}

\begin{table}[t]
\centering
\footnotesize
\renewcommand{\arraystretch}{1.12}
\setlength{\tabcolsep}{4pt}

\begin{tabularx}{\columnwidth}{>{\raggedright\arraybackslash}X >{\centering\arraybackslash}X}
\toprule
\textbf{Models} & \textbf{Macro-F1} \\
\midrule

LLaMA   & 0.90\std{0.76} \\
Gemma   & 0.89\std{1.34}\\
Mistral & 0.90\std{1.04}\\

\bottomrule
\end{tabularx}

\caption{Translation-based Macro-F1 performance of \textsc{SEEK} on \textsc{RU22Fact}.}
\label{tab:id_results}
\end{table}

\subsection{Findings on \textsc{RU22Fact}}
\noindent Table~\ref{tab:ru22fact_results} reports Macro-F1 results on the official \textsc{RU22Fact} test split. The results show that evidence quality strongly affects factual verification across different retrieval settings. The LLM-only setting is competitive, with LLaMA reaching \textbf{0.73} and Mistral reaching \textbf{0.72}, while sentence chunking provides only marginal gains. Semantic chunking improves performance, with LLaMA and Mistral both reaching \textbf{0.79}. However, \textsc{SEEK} gives the strongest results, with Gemma and Mistral achieving \textbf{0.90} and LLaMA reaching \textbf{0.89}. This gives a gain of \textbf{+0.17} over the best LLM-only result and \textbf{+0.11} over the best semantic chunking result. Overall, these results show that \textsc{SEEK} consistently improves fact verification by providing more complete and coherent evidence, enabling different LLMs to make more reliable predictions.

\subsection{Findings in Translation-based Retrieval}

\noindent We further evaluate translation-based retrieval with \textsc{SEEK}, where claims and crawled documents are translated into English before retrieval and prediction. As shown in Table~\ref{tab:cross_domain_results}, this setting performs strongly on \textsc{X-FACT}, with the best Macro-F1 scores of \textbf{0.66} in ID, \textbf{0.41} in OOD, and \textbf{0.37} in ZS. On \textsc{RU22Fact}, Table~\ref{tab:id_results} shows that all models achieve similarly high performance, with LLaMA and Mistral reaching \textbf{0.90} and Gemma reaching \textbf{0.89}. These results suggest that translation-based retrieval is complementary to \textsc{SEEK}: translation reduces linguistic variation, while \textsc{SEEK} preserves complete verification context.

\subsection{Evaluation of Retrieval Quality}

\noindent Beyond veracity prediction, we examine whether the retrieved evidence is useful for fact verification. Since gold evidence annotations are unavailable, we compare sentence chunking, semantic chunking, and \textsc{SEEK} using proxy analyses.

\begin{figure}[t]
    \centering
    \includegraphics[width=\columnwidth]{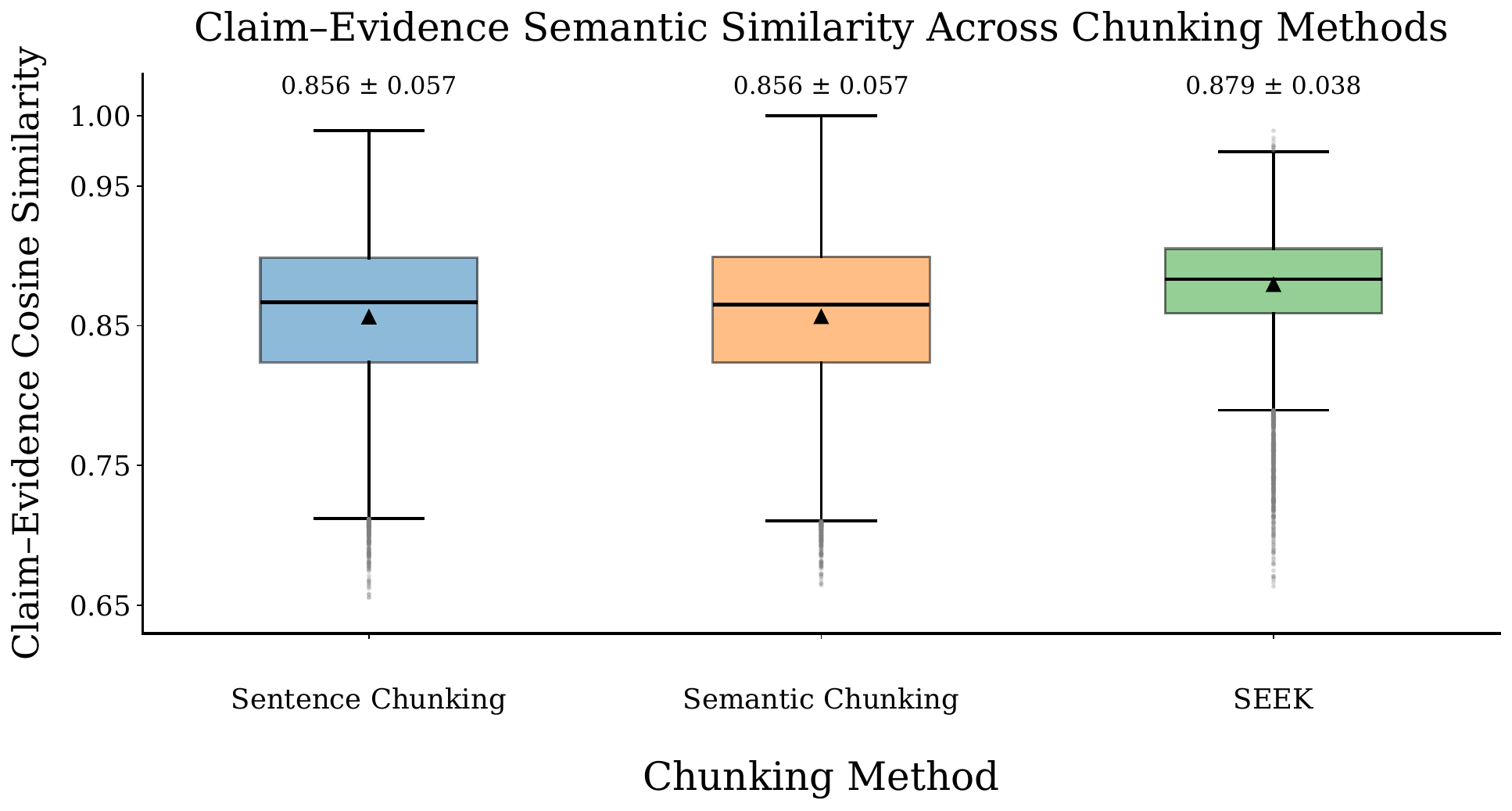}
    \caption{Claim-evidence semantic similarity across different chunking strategies.}
    \label{fig:xfact_claim_evidence_similarity}
\end{figure}

\noindent\textbf{Claim-Evidence Semantic Similarity:}
We compute cosine similarity between each claim and its retrieved evidence in the multilingual embedding space. Higher scores indicate stronger claim--evidence alignment, and Figure~\ref{fig:xfact_claim_evidence_similarity} shows the resulting distributions across chunking strategies.

\begin{figure}[t]
    \centering
    \includegraphics[width=\columnwidth]{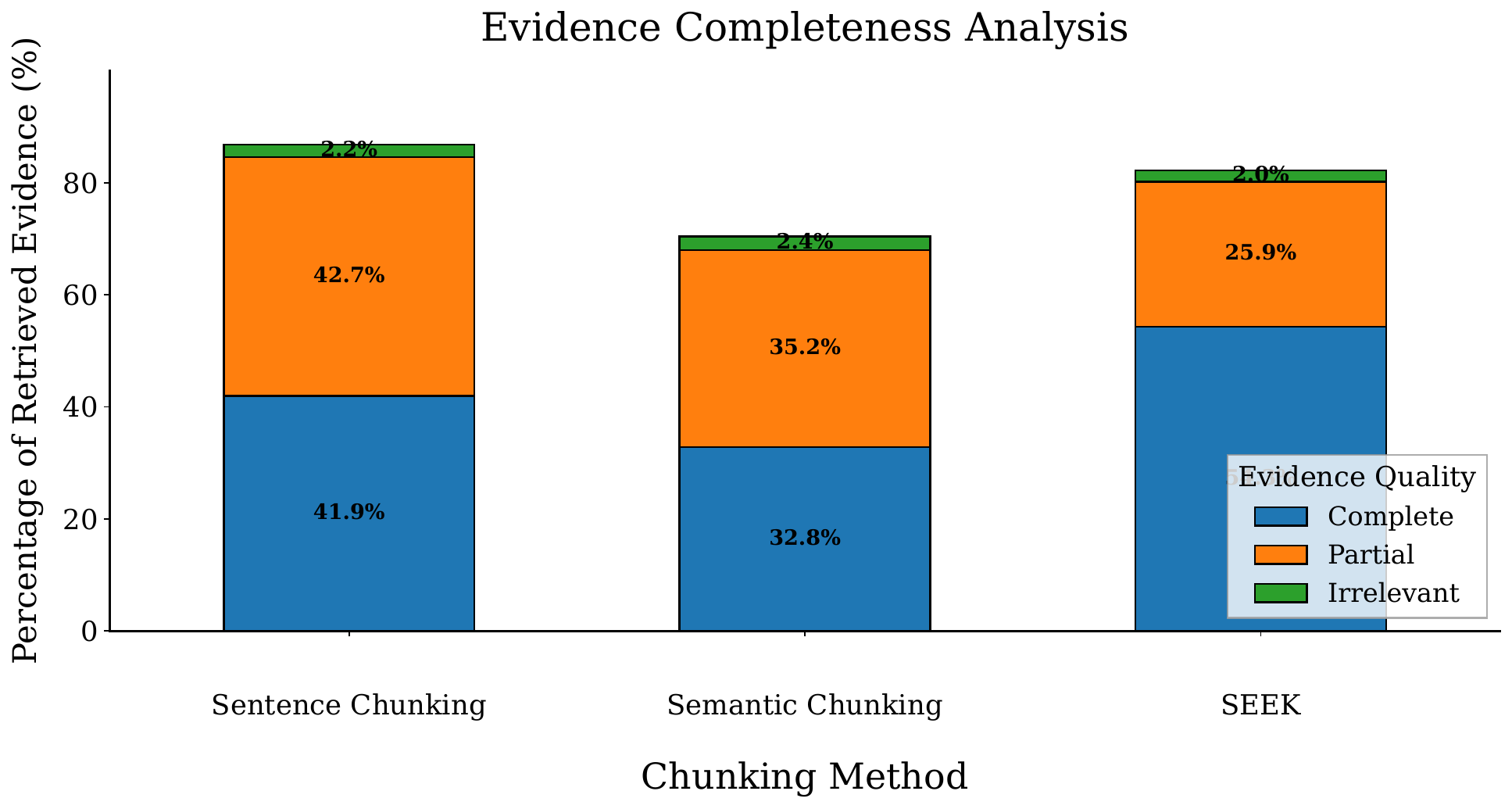}
    \caption{Evidence completeness analysis on X-FACT using an LLM-based evaluator.}
    \label{fig:xfact_evidence_completeness}
\end{figure}

\begin{figure}[t]
    \centering
    \includegraphics[width=\columnwidth]{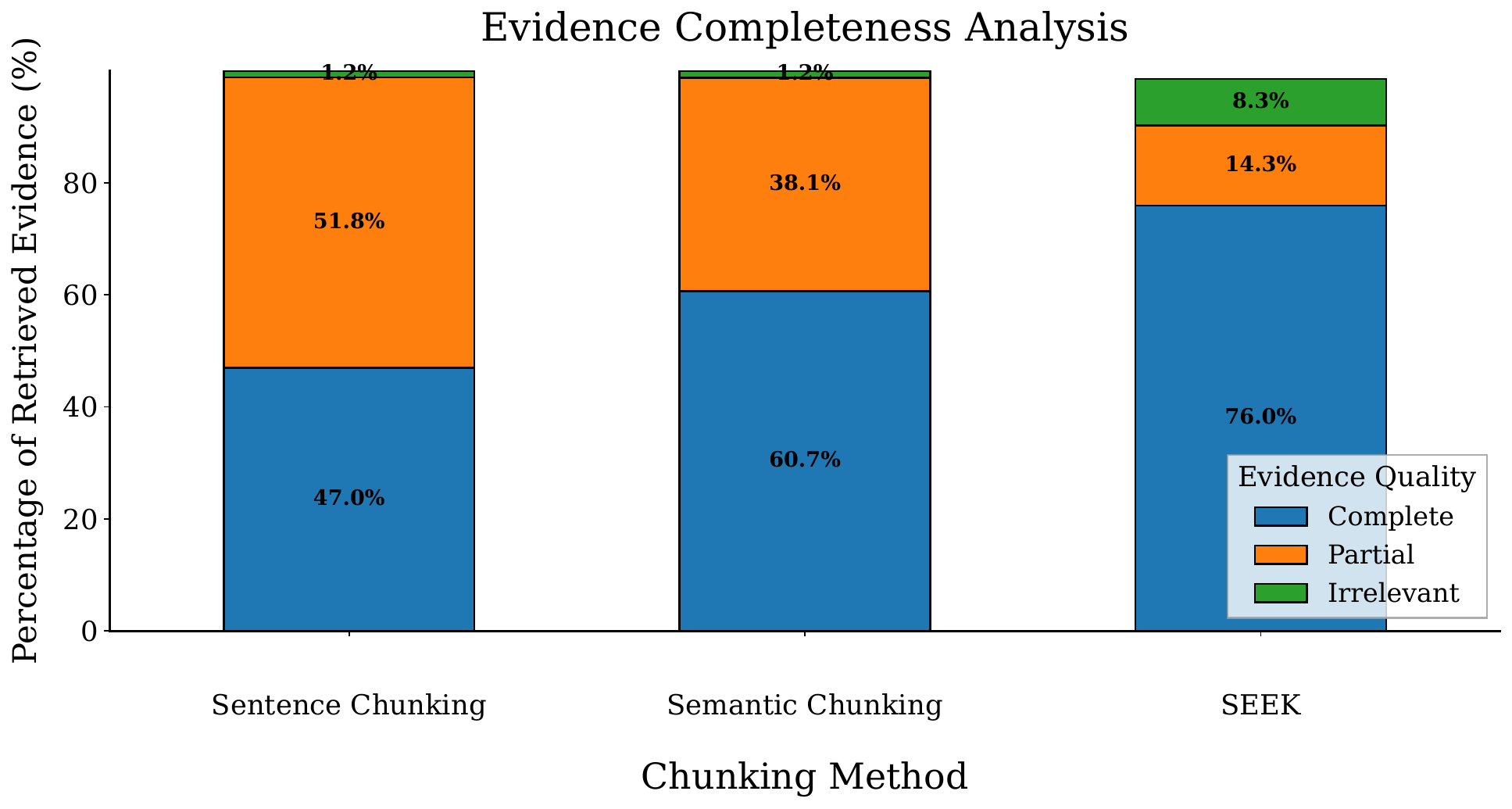}
    \caption{Evidence completeness analysis on \textsc{RU22Fact} using an LLM-based evaluator.}
    \label{fig:ru22fact_evidence_completeness}
\end{figure}

\noindent\textbf{Evidence Completeness Analysis:}
Because similarity alone does not ensure verification sufficiency, we use a locally deployed LLaMA-based evaluator to label each retrieved evidence instance as \textit{Complete}, \textit{Partial}, or \textit{Irrelevant}. The prompt is given in Appendix~\ref{app:evidence_completeness_prompt}. As shown in Figures~\ref{fig:xfact_evidence_completeness} and~\ref{fig:ru22fact_evidence_completeness}, \textsc{SEEK} retrieves more complete evidence than both baselines. Complete evidence reaches 54.3\% on \textsc{X-FACT}, compared with 41.9\% for sentence chunking and 32.8\% for semantic chunking. On \textsc{RU22Fact}, \textsc{SEEK} achieves 76.0\%, improving over sentence and semantic chunking by 29.0 and 15.3 percentage points, respectively, respectively. These results indicate that the proposed chunking strategy better preserves verification-ready context while reducing incomplete evidence retrieval.

\noindent\textbf{Qualitative and Significance Analysis:}
We further provide a qualitative comparison in Figure~\ref{fig:chunking_example}, focusing on semantic continuity, investigation flow, and whether the retrieved evidence contains final supporting or refuting statements. Additional examples are included in Appendix~\ref{app:chunking_example}. McNemar's test further confirms that \textsc{SEEK} yields statistically significant gains over other methods, particularly on \textsc{RU22Fact} and \textsc{X-FACT} ID. Significance tests are reported in Appendix~\ref{app:significance_tests}.

\section{Conclusion}

In this work, we proposed \textbf{SEEK}, a semantic evidence extraction framework for multilingual fact-checking. SEEK moves beyond short snippets, isolated sentences, and fixed-size chunks by building coherent evidence passages from full web documents. It detects topic shifts and keeps boundary context, allowing the retrieved evidence to better preserve the claim, supporting details, and final verification signal.
Experiments on X-FACT and \textsc{RU22Fact} show that SEEK improves veracity prediction across in-domain, out-of-domain, and zero-shot settings. The retrieval analysis further shows that SEEK retrieves evidence that is not only semantically relevant, but also more complete for verification.
Overall, our findings highlight that reliable multilingual fact-checking depends strongly on evidence construction. By providing focused and context-rich evidence chunks, SEEK helps downstream LLMs make more reliable factuality predictions and reduces the limitations of retrieval based on snippets, sentences, and standard semantic chunks.
\section{Limitations}

\textsc{SEEK} depends on the quality of crawled web documents; incomplete or noisy pages can still lead to missing evidence. Since the method detects semantic topic shifts, it may also miss cases where verification cues are spread across distant document sections. Our experiments are limited to \textsc{X-FACT} and \textsc{RU22Fact}, so evaluation on more languages, domains, and real-time fact-checking settings is needed. Finally, human-annotated evidence completeness labels would provide a stronger evaluation signal, which we leave for future work.


\bibliography{custom}

\appendix
\appendix
\section{Additional Details on Datasets}
\label{App:DatasetDetails}

This appendix provides additional details on the datasets used in our study, including their label distributions and evaluation splits.

\subsection{X-FACT }

X-FACT is a multilingual fact-checking dataset containing 31,189 claims collected from 85 fact-checking websites across 25 languages and 11 language families \cite{gupta2021xfact}. Each claim is annotated with a factuality label and includes metadata such as language, source URL, claim date, and review date.

The dataset contains seven labels: \textit{True, Mostly True, Partially True/Misleading, Mostly False, False, Complicated/Hard to categorize}, and \textit{Other}. As shown in Table~\ref{tab:xfact_label_distribution}.

For evaluation, X-FACT provides three test settings: \textit{In-domain}, where claims come from sources and languages seen during training; \textit{Out-of-domain}, where languages are seen but sources are unseen; and \textit{Zero-shot}, where both languages and sources are unseen during training. Table~\ref{tab:xfact_language_distribution} shows the language-wise distribution, highlighting that several languages appear only in the zero-shot split, making this setting useful for evaluating cross-lingual generalization.

\begin{table}[ht]
\centering
\small
\resizebox{\columnwidth}{!}{
\begin{tabular}{lrrrrr}
\toprule
\textbf{Class} & \textbf{Train} & \textbf{Dev} & \textbf{Indomain} & \textbf{OOD} & \textbf{Zeroshot} \\
\midrule
False & 7,515 & 977 & 1,464 & 1,109 & 1,471 \\
PT/M & 4,359 & 553 & 858 & 769 & 772 \\
True & 4,080 & 551 & 851 & 329 & 332 \\
MT & 1,380 & 194 & 300 & 0 & 73 \\
MF & 848 & 114 & 159 & 0 & 415 \\
C/H & 540 & 80 & 106 & 150 & 264 \\
Other & 357 & 66 & 88 & 11 & 54 \\
\midrule
\textbf{Total} & \textbf{19,079} & \textbf{2,535} & \textbf{3,826} & \textbf{2,368} & \textbf{3,381} \\
\bottomrule
\end{tabular}
}
\caption{Label-wise distribution of the X-FACT dataset across training, development, in-domain, out-of-domain, and zero-shot evaluation splits. PT/M = Partially True/Misleading, MT = Mostly True, MF = Mostly False, and C/H = Complicated/Hard to categorize.}
\label{tab:xfact_label_distribution}
\end{table}

\begin{table}[ht]
\centering
\scriptsize
\setlength{\tabcolsep}{3pt}
\resizebox{\columnwidth}{!}{
\begin{tabular}{lrrrrrr}
\toprule
\textbf{Language} & \textbf{Indomain} & \textbf{Dev} & \textbf{OOD} & \textbf{Train} & \textbf{Zeroshot} & \textbf{Total} \\
\midrule
Portuguese & 1121 & 747 & 472 & 5601 & 0 & 7941 \\
Indonesian & 448 & 297 & 647 & 2231 & 0 & 3623 \\
Arabic & 314 & 209 & 0 & 1567 & 0 & 2090 \\
Georgian & 307 & 203 & 0 & 1529 & 0 & 2039 \\
Polish & 267 & 176 & 0 & 1325 & 0 & 1768 \\
Turkish & 169 & 111 & 613 & 858 & 0 & 1751 \\
Italian & 190 & 125 & 255 & 943 & 0 & 1513 \\
Hindi & 170 & 112 & 381 & 845 & 0 & 1508 \\
Tamil & 221 & 145 & 0 & 1097 & 0 & 1463 \\
Spanish & 211 & 139 & 0 & 1049 & 0 & 1399 \\
German & 142 & 95 & 0 & 712 & 0 & 949 \\
Romanian & 140 & 93 & 0 & 698 & 0 & 931 \\
Serbian & 126 & 83 & 0 & 624 & 0 & 833 \\
Albanian & 0 & 0 & 0 & 0 & 666 & 666 \\
Bengali & 0 & 0 & 0 & 0 & 568 & 568 \\
Russian & 0 & 0 & 0 & 0 & 442 & 442 \\
Norwegian & 0 & 0 & 0 & 0 & 406 & 406 \\
Persian & 0 & 0 & 0 & 0 & 281 & 281 \\
Azerbaijani & 0 & 0 & 0 & 0 & 253 & 253 \\
French & 0 & 0 & 0 & 0 & 198 & 198 \\
Dutch & 0 & 0 & 0 & 0 & 166 & 166 \\
Gujarati & 0 & 0 & 0 & 0 & 129 & 129 \\
Punjabi & 0 & 0 & 0 & 0 & 105 & 105 \\
Marathi & 0 & 0 & 0 & 0 & 87 & 87 \\
Sinhala & 0 & 0 & 0 & 0 & 80 & 80 \\
\midrule
\textbf{Total} & \textbf{3826} & \textbf{2535} & \textbf{2368} & \textbf{19079} & \textbf{3381} & \textbf{31189} \\
\bottomrule
\end{tabular}
}
\caption{Language-wise distribution of the X-FACT dataset across evaluation splits. Languages appearing only in the zero-shot split are unseen during training and are used to evaluate cross-lingual transfer.}
\label{tab:xfact_language_distribution}
\end{table}

\subsection{\textsc{RU22Fact}}

\textsc{RU22Fact} is a multilingual explainable fact-checking dataset focused on the 2022 Russia--Ukraine conflict \cite{zeng2024ru22fact}. It contains 16,033 claims in four languages: English, Chinese, Russian, and Ukrainian. Each sample includes a claim, optimized evidence, a reference explanation, metadata, and one of three labels: \textit{Supported}, \textit{Refuted}, or \textit{Not Enough Information} (NEI).

Table~\ref{tab:ru22fact_label_distribution} shows the label distribution, where \textit{Supported} is the largest class. Table~\ref{tab:ru22fact_language_distribution} reports the language-wise distribution, with English forming the largest portion of the dataset.

\begin{table}[ht]
\centering
\small
\resizebox{\columnwidth}{!}{
\begin{tabular}{lrrrr}
\toprule
\textbf{Class} & \textbf{Train} & \textbf{Dev} & \textbf{Test} & \textbf{Total} \\
\midrule
Supported & 6,836 & 1,079 & 2,166 & 10,081 \\
Refuted   & 3,299 & 450   & 902   & 4,651 \\
NEI       & 1,082 & 71    & 148   & 1,301 \\
\midrule
\textbf{Total} & \textbf{11,217} & \textbf{1,600} & \textbf{3,216} & \textbf{16,033} \\
\bottomrule
\end{tabular}
}
\caption{Label-wise distribution of the \textsc{RU22Fact} dataset across training, development, and test splits. NEI denotes Not Enough Information.}
\label{tab:ru22fact_label_distribution}
\end{table}

\begin{table}[ht]
\centering
\small
\resizebox{\columnwidth}{!}{
\begin{tabular}{lrrrr}
\toprule
\textbf{Language} & \textbf{Train} & \textbf{Dev} & \textbf{Test} & \textbf{Total} \\
\midrule
English    & 6,082 & 867 & 1,741 & 8,690 \\
Chinese    & 1,055 & 152 & 305   & 1,512 \\
Russian    & 1,621 & 231 & 465   & 3,399 \\
Ukrainian  & 2,458 & 350 & 704   & 3,512 \\
\midrule
\textbf{Total} & \textbf{11,217} & \textbf{1,600} & \textbf{3,216} & \textbf{16,033} \\
\bottomrule
\end{tabular}
}
\caption{Language-wise distribution of the \textsc{RU22Fact} dataset across training, development, and test splits.}
\label{tab:ru22fact_language_distribution}
\end{table}

\section{Qualitative Analysis of Chunk Formation}
\label{app:chunking_example}

\begin{figure}[ht]
    \centering
    \includegraphics[width=\columnwidth]{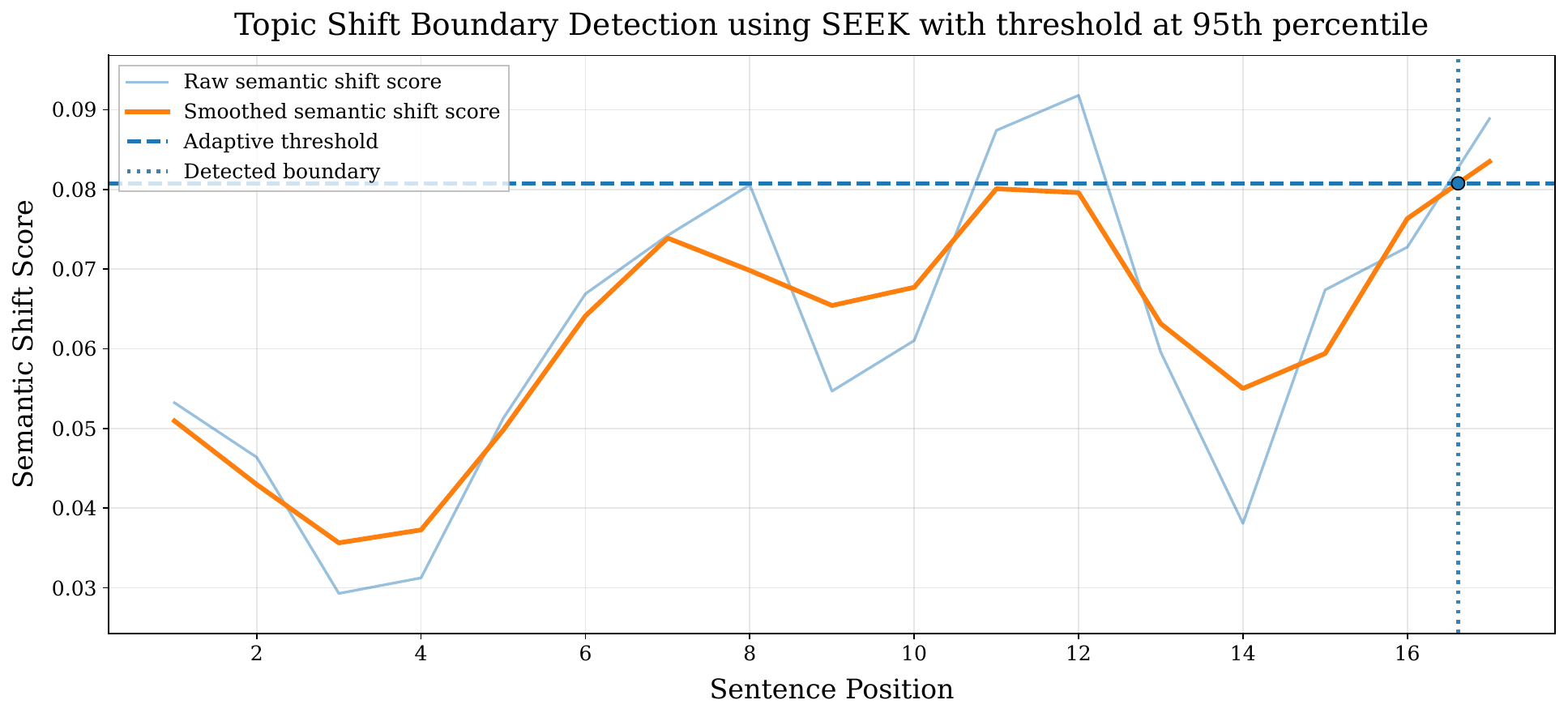}
    \caption{Example of topic-shift boundary detection using SEEK on a Hindi fact-checking document. The smoothed semantic shift curve reduces local sentence-level noise, while the adaptive threshold selects meaningful chunk boundaries.}
    \label{fig:topic_shift_doc1}
\end{figure}

\begin{figure*}[ht]
    \centering
    \includegraphics[width=\textwidth]{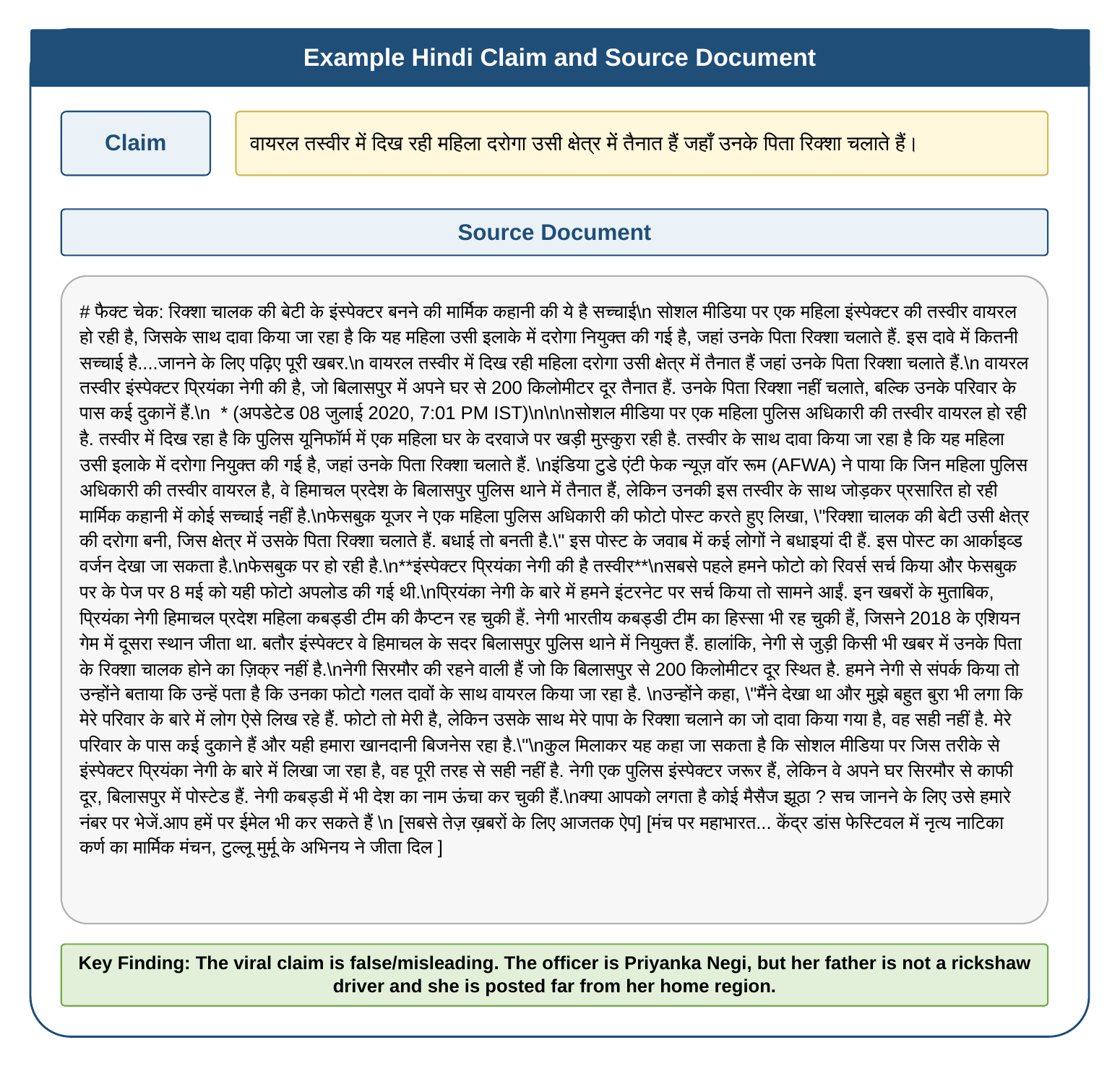}
    \caption{Example Hindi claim and source document.}
    \label{fig:example_claim_document}
\end{figure*}

To qualitatively illustrate the proposed Semantic Evidence Extraction with adaptive chunKing (\textsc{SEEK}) method, we use a real Hindi fact-checking example consisting of a viral claim and its corresponding source article, as shown in Figure~\ref{fig:example_claim_document}. The claim states that the woman inspector in the viral image is posted in the same area where her father works as a rickshaw driver. The source article investigates this claim and concludes that the viral story is misleading. Figure~\ref{fig:topic_shift_doc1} shows the raw semantic shift scores, their smoothed version, and the adaptive threshold used by \textsc{SEEK} to identify meaningful topic boundaries.

\begin{figure*}[ht]
    \centering
        \includegraphics[width=\textwidth]{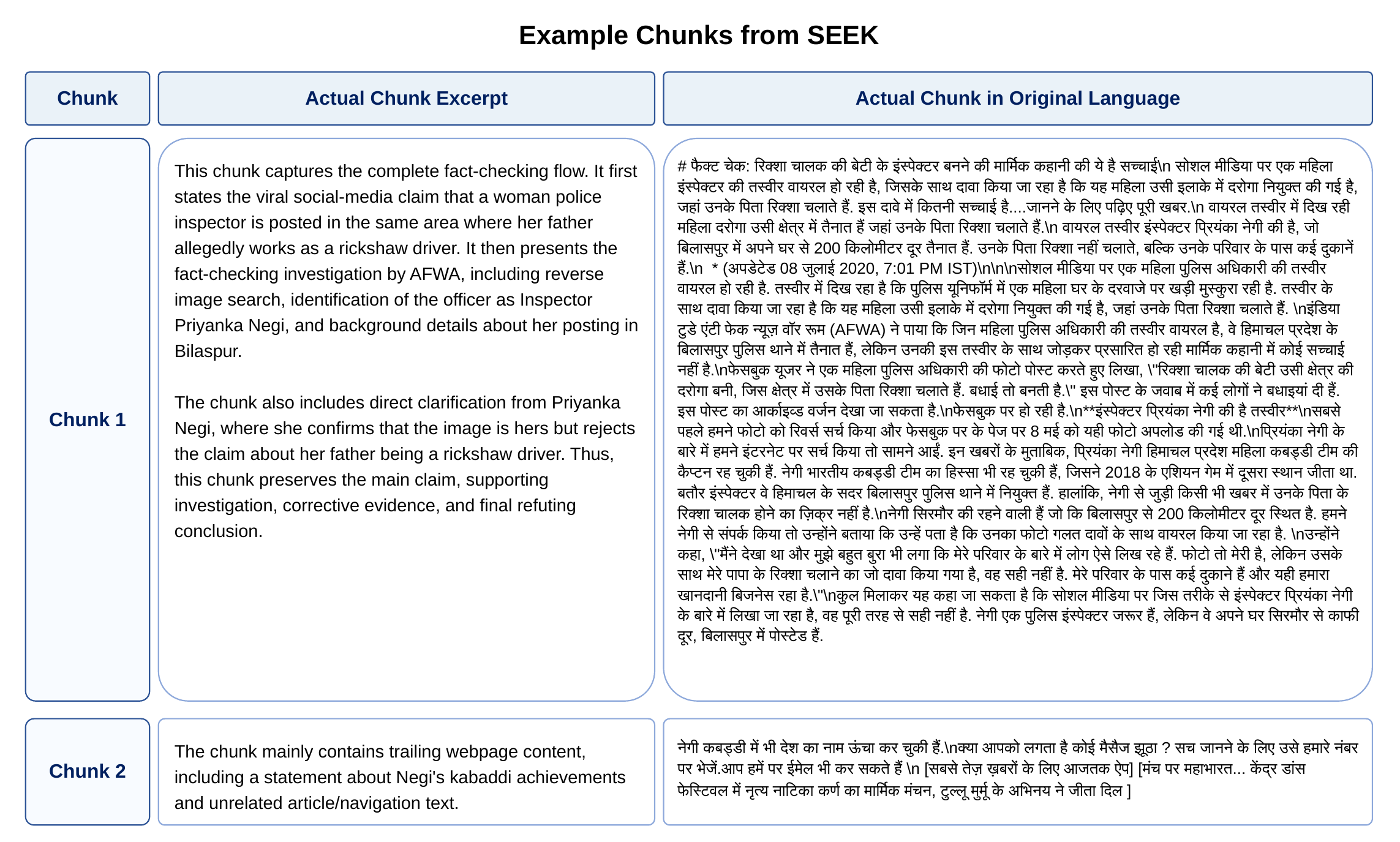}
    \caption{Example chunks produced by the proposed Semantic Evidence Extraction with Adaptive chunKing(SEEK) method on a Hindi fact-checking document. The figure reports English summaries and original Hindi chunks}
    \label{fig:Improved_semantic_chunking_example_document}
\end{figure*}

\begin{figure*}[ht]
    \centering
        \includegraphics[width=\textwidth]{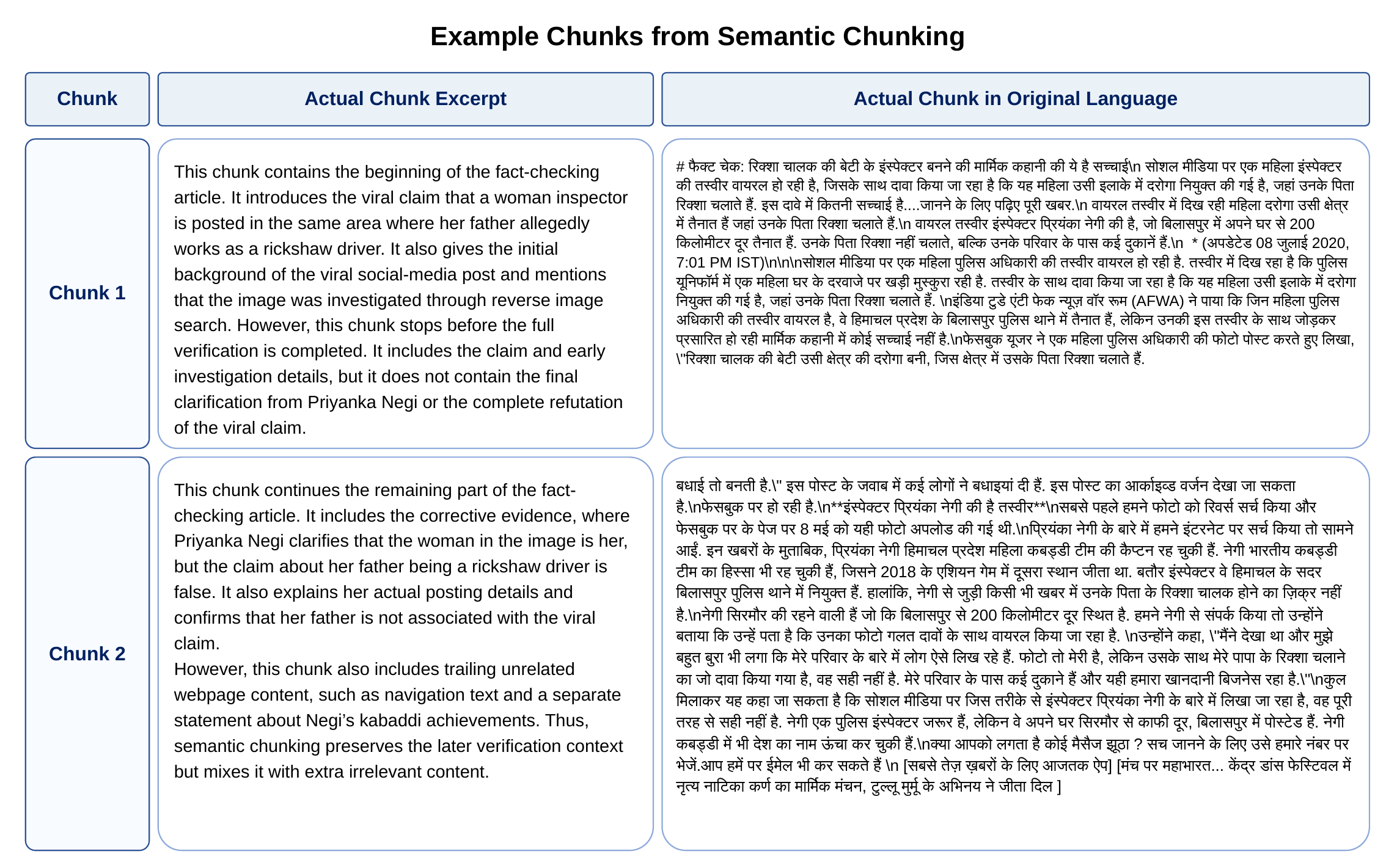}
    \caption{Example chunks produced by the semantic chunking method on a Hindi fact-checking document. The figure reports English summaries and original Hindi chunks}
    \label{fig:semantic_chunking_example_document}
\end{figure*}

\begin{figure*}[ht]
    \centering
        \includegraphics[width=\textwidth]{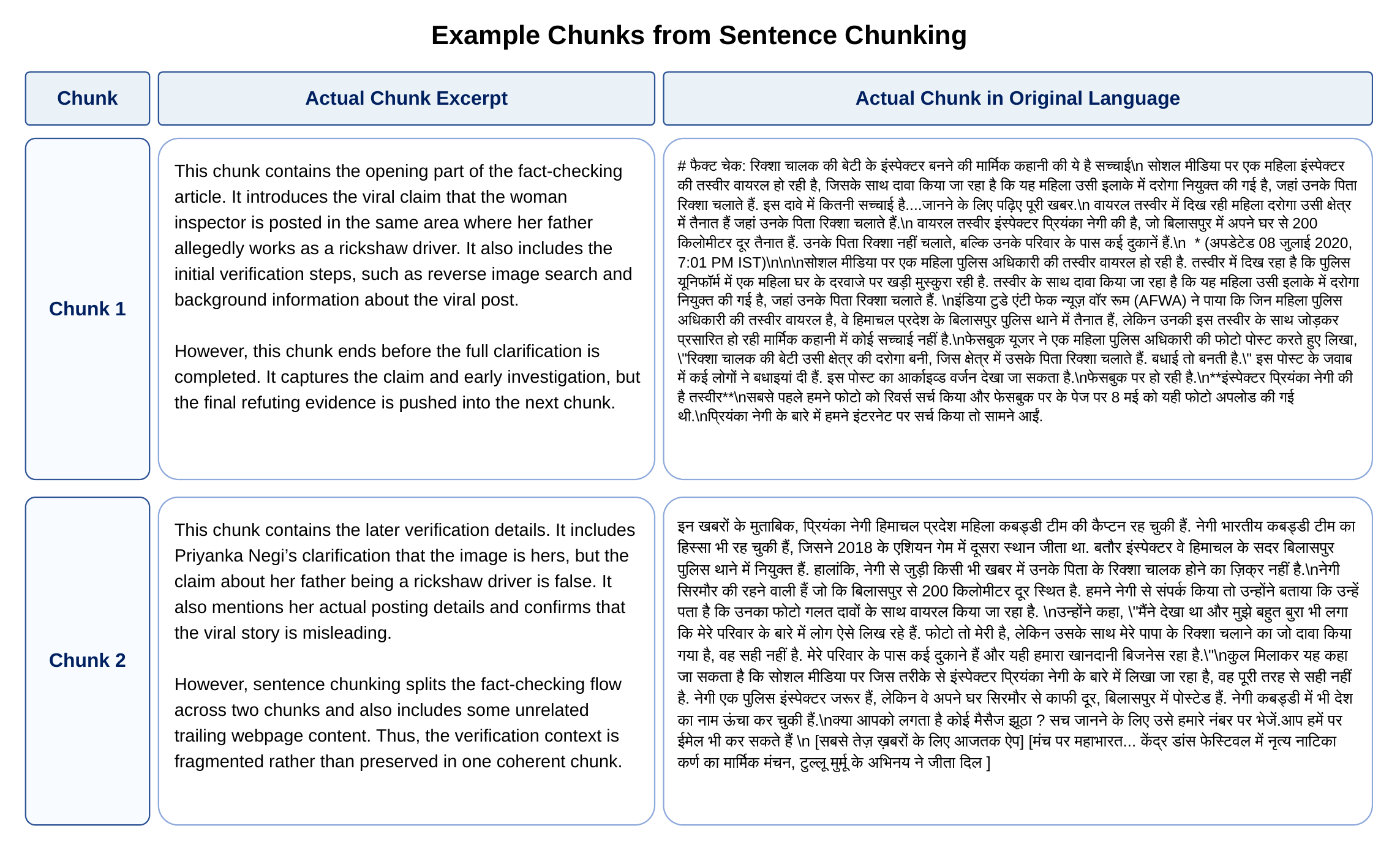}
    \caption{Example chunks produced by the sentence chunking method on a Hindi fact-checking document. The figure reports English summaries and original Hindi chunks}
    \label{fig:sentence_chunking_example_document}
\end{figure*}

\noindent Figure~\ref{fig:Improved_semantic_chunking_example_document} shows the chunks produced by the proposed \textsc{SEEK} method. The first chunk preserves the complete fact-checking flow, including the viral claim, evidence investigation, corrective clarification, and final refutation, while the second chunk mainly contains trailing unrelated webpage content. In contrast, Figure~\ref{fig:semantic_chunking_example_document} shows that semantic chunking splits the verification context across two chunks: the first chunk contains the claim and early investigation, whereas the second chunk contains the final clarification but also mixes it with unrelated navigation content. Similarly, Figure~\ref{fig:sentence_chunking_example_document} shows that sentence chunking fragments the fact-checking flow into separate sentence-budgeted chunks, pushing the decisive refuting evidence into the second chunk and reducing evidence completeness within a single retrieved chunk. This example demonstrates that \textsc{SEEK} better separates the core verification evidence from less relevant trailing content.

\section{Dataset-Specific Instructions}
\label{app:instructions}

We use dataset-specific task instructions to guide the language model for veracity prediction. The instruction is prepended to the claim and retrieved evidence chunks, and the exact prompts used for X-FACT and \textsc{RU22Fact} are reported in Table~\ref{tab:dataset_specific_instructions}.

\begin{table*}[ht]
\centering
\small
\setlength{\tabcolsep}{5pt}
\renewcommand{\arraystretch}{1.25}
\begin{tabular}{p{0.16\textwidth} p{0.84\textwidth}}
\toprule
\textbf{Dataset} & \textbf{Instruction} \\
\midrule

X-FACT 
& Classify the given \{language\} claim into one of the seven categories: 
\textit{TRUE, MOSTLY-TRUE, PARTLY-TRUE/MISLEADING, FALSE, MOSTLY-FALSE, COMPLICATED/HARD-TO-CATEGORISE, OTHER}, based on the provided evidence. 
The label definitions are: TRUE = fully supported by the evidence; MOSTLY-TRUE = mostly supported but with minor inaccuracies; PARTLY-TRUE/MISLEADING = partially supported but with significant omissions; FALSE = clearly contradicted or unsupported; MOSTLY-FALSE = largely incorrect with only a small element of truth; COMPLICATED/HARD-TO-CATEGORISE = too complex for a straightforward label; OTHER = does not fit the above categories. 
Provide exactly one label. \\

\midrule

\textsc{RU22Fact}
& Classify the given \{language\} claim into one of three categories: 
\textit{SUPPORTED, REFUTED, NEI}, based on the provided evidence. 
The label definitions are: SUPPORTED = the evidence supports the claim; REFUTED = the evidence contradicts the claim; NEI = not enough information to verify. 
Provide exactly one label. \\

\bottomrule
\end{tabular}
\caption{Dataset-specific task instructions used for veracity prediction.}
\label{tab:dataset_specific_instructions}
\end{table*}

\section{Model Details}
\label{app:model_details}

Table~\ref{tab:model_details} summarizes the encoder and decoder models used in our framework. The encoder models are used for multilingual claim--evidence representation and retrieval, while the instruction-tuned large language models are used for downstream veracity prediction.

\begin{table*}[t]
\centering
\small
\setlength{\tabcolsep}{27.5pt}
\renewcommand{\arraystretch}{1.05}
\begin{tabular}{llll}
\toprule
\textbf{Component} & \textbf{Model Family} & \textbf{Model} & \textbf{Size} \\
\midrule
Retrieval encoder & Multilingual E5 & multilingual-e5-large-instruct & 560M \\
Veracity prediction & LLaMA & LLaMA-3.1-8B-Instruct & 8B \\
Veracity prediction & Gemma & Gemma-2-9B-Instruct & 9B \\
Veracity prediction & Mistral & Mistral-7B-Instruct-v0.3 & 7B \\
\bottomrule
\end{tabular}
\caption{Details of encoder and decoder models used in our experiments.}
\label{tab:model_details}
\end{table*}

\noindent We use multilingual-e5-large as the retrieval encoder because our task involves fact-checking across multiple languages. This encoder provides a shared multilingual representation space, allowing claims and evidence chunks within each language to be encoded consistently for dense retrieval. FAISS is used to perform efficient nearest-neighbour search over the global chunk pool.

\noindent For veracity prediction, we evaluate three instruction-tuned LLM families: LLaMA, Gemma, and Mistral. These models are selected because they are strong open-source multilingual instruction-following models with different architectures and training strategies. Evaluating multiple model families allows us to examine whether the proposed chunking strategy consistently improves downstream fact verification performance, rather than being effective only for a single model.
\subsection{Hyperparameters}
\label{app:hyperparameters}

All veracity prediction models are fine-tuned for \texttt{3} epochs using LoRA adaptation. Unless otherwise specified, the LoRA rank is set to \texttt{8}. The same retrieval configuration, decoding settings, and evaluation protocol are used across all experiments to ensure a fair comparison between different chunking strategies.

\section{Prompt for Evidence Completeness Evaluation}
\label{app:evidence_completeness_prompt}

We use the prompt shown in Table~\ref{tab:evidence_completeness_prompt} to evaluate whether the retrieved evidence contains sufficient verification context.

\begin{table*}[t]
\centering

\begin{minipage}[t]{0.40\textwidth}
\centering
\small
\setlength{\tabcolsep}{8pt}
\renewcommand{\arraystretch}{1.05}
\begin{tabular}{ll}
\toprule
\textbf{Hyperparameter} & \textbf{Value} \\
\midrule
Fine-tuning method & LoRA \\
Number of epochs & \texttt{3} \\
LoRA rank & \texttt{8} \\
Retrieved candidates & \texttt{20} \\
Final evidence chunks & \texttt{5} \\
Retrieval method & FAISS dense retrieval \\
Evaluation metric & Macro-F1 \\
\bottomrule
\end{tabular}
\caption{Hyperparameter settings used across experiments.}
\label{tab:hyperparameters}
\end{minipage}
\hfill
\begin{minipage}[t]{0.56\textwidth}
\centering
\scriptsize
\renewcommand{\arraystretch}{1.05}
\begin{tabular}{p{\linewidth}}
\toprule
You are evaluating retrieved evidence for a fact-checking task. \\[0.3em]

\textbf{Claim:} \{claim\} \\[0.3em]

\textbf{Retrieved Evidence:} \{evidence\} \\[0.3em]

\textbf{Question:} Does the retrieved evidence contain enough information to verify whether the claim is true or false? \\[0.3em]

Choose exactly one label: \\[0.1em]

\textbf{Complete:} The evidence contains sufficient information to verify the claim. \\

\textbf{Partial:} The evidence is related to the claim but incomplete. \\

\textbf{Irrelevant:} The evidence is unrelated, noisy, empty, blocked webpage text, or does not help verify the claim. \\[0.3em]

Return only one word: \textit{Complete}, \textit{Partial}, or \textit{Irrelevant}. \\
\bottomrule
\end{tabular}
\caption{Prompt used for evidence completeness evaluation.}
\label{tab:evidence_completeness_prompt}
\end{minipage}

\end{table*}

\begin{figure*}[ht!]
    \centering
    \includegraphics[width=\linewidth]{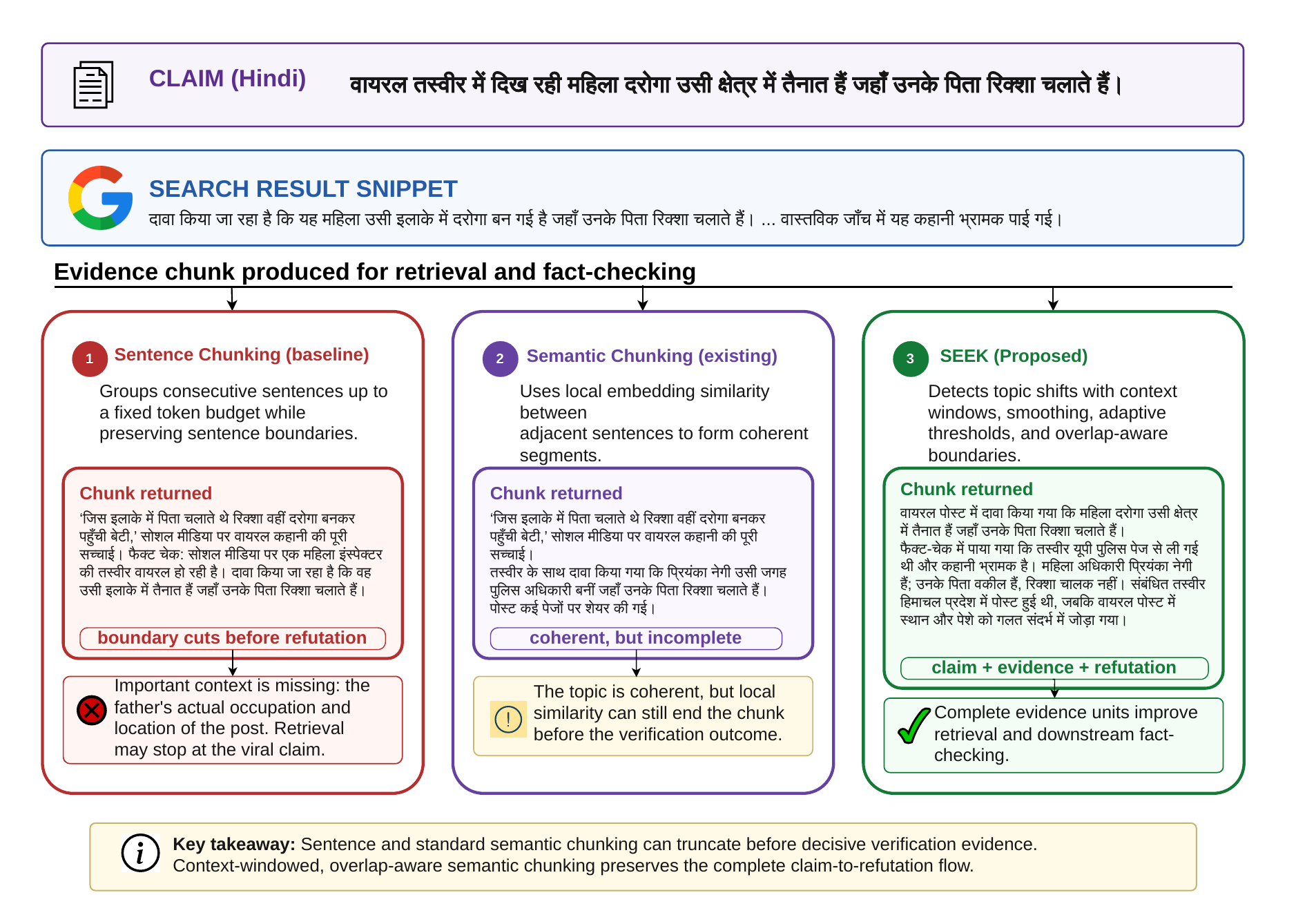}
    \caption{Illustration of the chunking strategies used for evidence retrieval. Sentence chunking groups consecutive sentences under a token budget but may truncate the evidence before the full verification context is reached. Existing semantic chunking uses embedding-based similarity to form coherent chunks, but can still miss later refuting evidence in noisy web documents. Our SEEK preserves a longer and more complete evidence unit by detecting topic shifts using contextual windows, smoothing, adaptive thresholding, and boundary overlap.}
    \label{fig:chunking_example}
\end{figure*}

\begin{table*}[t]
\centering
\footnotesize
\renewcommand{\arraystretch}{1.22}
\setlength{\tabcolsep}{7pt}

\definecolor{lightgreen}{HTML}{EAF7E8}
\definecolor{lightgray}{HTML}{F5F5F5}

\begin{tabular}{l l l c c c c c c}
\toprule
\textbf{Dataset} & \textbf{Split} & \textbf{Model} 
& \textbf{B$>$O} & \textbf{O$>$B} 
& \textbf{Net Gain} 
& \textbf{McNemar $\chi^2$} 
& \textbf{$p$-value} 
& \textbf{Significance} \\
\midrule

\multirow{3}{*}{\textsc{RU22Fact}} 
& Test & Gemma   & 31  & 136 & +105 & 64.77 & $8.88{\times}10^{-16}$ & \cellcolor{lightgreen}\checkmark \\

& Test & LLaMA   & 41  & 147 & +106 & 58.64 & $1.89{\times}10^{-14}$ & \cellcolor{lightgreen}\checkmark \\

& Test & Mistral & 41  & 144 & +103 & 56.24 & $6.42{\times}10^{-14}$ & \cellcolor{lightgreen}\checkmark \\

\midrule
\multirow{9}{*}{\textsc{X-FACT}} 
& ID  & Gemma   & 306 & 487 & +181 & 40.86 & $1.64{\times}10^{-10}$ & \cellcolor{lightgreen}\checkmark \\
& ID  & LLaMA   & 255 & 410 & +155 & 35.66 & $2.35{\times}10^{-9}$  & \cellcolor{lightgreen}\checkmark \\
& ID  & Mistral & 290 & 497 & +207 & 53.92 & $2.09{\times}10^{-13}$ & \cellcolor{lightgreen}\checkmark \\

\cmidrule(lr){2-9}

& OOD & Gemma   & 187 & 329 & +142 & 38.53 & $5.39{\times}10^{-10}$ & \cellcolor{lightgreen}\checkmark \\
& OOD & LLaMA   & 171 & 298 & +127 & 33.85 & $5.95{\times}10^{-9}$  & \cellcolor{lightgreen}\checkmark \\
& OOD & Mistral & 198 & 229 & +31  & 2.11  & $1.47{\times}10^{-1}$  & $\times$ \\

\cmidrule(lr){2-9}

& ZS  & Gemma   & 303 & 371 & +68  & 6.66  & $9.86{\times}10^{-3}$ & \cellcolor{lightgreen}\checkmark \\
& ZS  & LLaMA   & 273 & 394 & +121 & 21.59 & $3.38{\times}10^{-6}$ & \cellcolor{lightgreen}\checkmark \\
& ZS  & Mistral & 322 & 389 & +67  & 6.13  & $1.33{\times}10^{-2}$ & \cellcolor{lightgreen}\checkmark \\

\bottomrule
\end{tabular}

\caption{McNemar significance test comparing our SEEK method against standard semantic chunking. B$>$O denotes examples correctly predicted only by the baseline, while O$>$B denotes examples correctly predicted only by our method. Net Gain is computed as O$>$B minus B$>$O. Significance is reported at $\alpha=0.05$.}
\label{tab:mcnemar_ours_vs_semantic}
\end{table*}
\section{Significance Test Results}
\label{app:significance_tests}

We report all McNemar significance test results in this appendix to assess whether the gains of \textsc{SEEK} over different evidence construction baselines are statistically reliable. Table~\ref{tab:mcnemar_ours_vs_semantic} compares \textsc{SEEK} with semantic chunking, Table~\ref{tab:appendix_mcnemar_ours_vs_sentence} compares it with sentence chunking, Table~\ref{tab:appendix_mcnemar_ours_vs_concrete} compares it with the \textsc{CONCRETE} baseline, Table~\ref{tab:appendix_mcnemar_ours_vs_google_snippet} compares it with Google Search snippets, and Table~\ref{tab:appendix_mcnemar_ours_vs_llm} compares it with the LLM-only baseline. Across all tables, B$>$O denotes examples correctly predicted only by the baseline, whereas O$>$B denotes examples correctly predicted only by \textsc{SEEK}. Net Gain is computed as O$>$B minus B$>$O, and statistical significance is reported at $\alpha=0.05$.
\begin{table*}[t]
\centering
\footnotesize
\renewcommand{\arraystretch}{1.18}
\setlength{\tabcolsep}{6pt}

\begin{tabular}{l l l c c c c c c}
\toprule
\textbf{Dataset} & \textbf{Split} & \textbf{Model}
& \textbf{B$>$O} & \textbf{O$>$B}
& \textbf{Net Gain}
& \textbf{McNemar $\chi^2$}
& \textbf{$p$-value}
& \textbf{Significance} \\
\midrule

\multirow{3}{*}{\textsc{RU22Fact}}
& Test & Gemma   & 33 & 283 & +250 & 196.21 & $<10^{-16}$ & \cellcolor{lightgreen}\checkmark \\
& Test & LLaMA   & 37 & 297 & +260 & 200.84 & $<10^{-16}$ & \cellcolor{lightgreen}\checkmark \\
& Test & Mistral & 32 & 294 & +262 & 208.96 & $<10^{-16}$ & \cellcolor{lightgreen}\checkmark \\

\midrule

\multirow{9}{*}{\textsc{X-FACT}}
& ID  & Gemma   & 310 & 465 & +155 & 30.60 & $3.17{\times}10^{-8}$  & \cellcolor{lightgreen}\checkmark \\
& ID  & LLaMA   & 273 & 433 & +160 & 35.81 & $2.18{\times}10^{-9}$  & \cellcolor{lightgreen}\checkmark \\
& ID  & Mistral & 283 & 408 & +125 & 22.25 & $2.39{\times}10^{-6}$  & \cellcolor{lightgreen}\checkmark \\

\cmidrule(lr){2-9}

& OOD & Gemma   & 212 & 344 & +132 & 30.87 & $2.77{\times}10^{-8}$  & \cellcolor{lightgreen}\checkmark \\
& OOD & LLaMA   & 208 & 241 & +33  & 2.28  & $1.31{\times}10^{-1}$  & $\times$ \\
& OOD & Mistral & 158 & 297 & +139 & 41.85 & $9.83{\times}10^{-11}$ & \cellcolor{lightgreen}\checkmark \\

\cmidrule(lr){2-9}

& ZS  & Gemma   & 364 & 367 & +3   & 0.01  & $9.41{\times}10^{-1}$  & $\times$ \\
& ZS  & LLaMA   & 259 & 402 & +143 & 30.51 & $3.33{\times}10^{-8}$  & \cellcolor{lightgreen}\checkmark \\
& ZS  & Mistral & 347 & 465 & +118 & 16.86 & $4.03{\times}10^{-5}$  & \cellcolor{lightgreen}\checkmark \\

\bottomrule
\end{tabular}

\caption{McNemar significance test comparing our SEEK method against sentence chunking.}
\label{tab:appendix_mcnemar_ours_vs_sentence}
\end{table*}

\begin{table*}[t]
\centering
\footnotesize
\renewcommand{\arraystretch}{1.18}
\setlength{\tabcolsep}{6pt}

\definecolor{lightgreen}{HTML}{EAF7E8}

\begin{tabular}{l l l c c c c c c}
\toprule
\textbf{Dataset} & \textbf{Split} & \textbf{Model}
& \textbf{B$>$O} & \textbf{O$>$B}
& \textbf{Net Gain}
& \textbf{McNemar $\chi^2$}
& \textbf{$p$-value}
& \textbf{Significance} \\
\midrule

\multirow{9}{*}{\textsc{X-FACT}}
& ID  & Gemma   & 294 & 1225 & +931 & 569.39 & $<10^{-16}$ & \cellcolor{lightgreen}\checkmark \\
& ID  & LLaMA   & 260 & 1084 & +824 & 503.97 & $<10^{-16}$ & \cellcolor{lightgreen}\checkmark \\
& ID  & Mistral & 290 & 1186 & +896 & 542.70 & $<10^{-16}$ & \cellcolor{lightgreen}\checkmark \\

\cmidrule(lr){2-9}

& OOD & Gemma   & 235 & 563 & +328 & 134.00 & $<10^{-16}$ & \cellcolor{lightgreen}\checkmark \\
& OOD & LLaMA   & 199 & 568 & +369 & 176.56 & $<10^{-16}$ & \cellcolor{lightgreen}\checkmark \\
& OOD & Mistral & 180 & 588 & +408 & 215.69 & $<10^{-16}$ & \cellcolor{lightgreen}\checkmark \\

\cmidrule(lr){2-9}

& ZS  & Gemma   & 357 & 734 & +377 & 129.58 & $<10^{-16}$ & \cellcolor{lightgreen}\checkmark \\
& ZS  & LLaMA   & 337 & 721 & +384 & 138.65 & $<10^{-16}$ & \cellcolor{lightgreen}\checkmark \\
& ZS  & Mistral & 365 & 669 & +304 & 88.79  & $<10^{-16}$ & \cellcolor{lightgreen}\checkmark \\

\bottomrule
\end{tabular}

\caption{McNemar significance test comparing our SEEK method against the concrete baseline.}
\label{tab:appendix_mcnemar_ours_vs_concrete}

\end{table*}

\begin{table*}[t]
\centering
\footnotesize
\renewcommand{\arraystretch}{1.18}
\setlength{\tabcolsep}{6pt}

\begin{tabular}{l l l c c c c c c}
\toprule
\textbf{Dataset} & \textbf{Split} & \textbf{Model}
& \textbf{B$>$O} & \textbf{O$>$B}
& \textbf{Net Gain}
& \textbf{McNemar $\chi^2$}
& \textbf{$p$-value}
& \textbf{Significance} \\
\midrule

\multirow{9}{*}{\textsc{X-FACT}}
& ID  & Gemma   & 356 & 1052 & +696 & 343.06 & $<10^{-16}$ & \cellcolor{lightgreen}\checkmark \\
& ID  & LLaMA   & 320 & 1145 & +825 & 463.46 & $<10^{-16}$ & \cellcolor{lightgreen}\checkmark \\
& ID  & Mistral & 359 & 1031 & +672 & 323.91 & $<10^{-16}$ & \cellcolor{lightgreen}\checkmark \\

\cmidrule(lr){2-9}

& OOD & Gemma   & 227 & 503 & +276 & 103.60 & $<10^{-16}$ & \cellcolor{lightgreen}\checkmark \\
& OOD & LLaMA   & 246 & 522 & +276 & 98.47  & $<10^{-16}$ & \cellcolor{lightgreen}\checkmark \\
& OOD & Mistral & 184 & 466 & +282 & 121.48 & $<10^{-16}$ & \cellcolor{lightgreen}\checkmark \\

\cmidrule(lr){2-9}

& ZS  & Gemma   & 417 & 515 & +98  & 10.10 & $1.49{\times}10^{-3}$ & \cellcolor{lightgreen}\checkmark \\
& ZS  & LLaMA   & 349 & 646 & +297 & 88.06 & $<10^{-16}$ & \cellcolor{lightgreen}\checkmark \\
& ZS  & Mistral & 364 & 622 & +258 & 66.99 & $2.22{\times}10^{-16}$ & \cellcolor{lightgreen}\checkmark \\

\bottomrule
\end{tabular}

\caption{McNemar significance test comparing our SEEK method against the Google snippet baseline.}
\label{tab:appendix_mcnemar_ours_vs_google_snippet}

\end{table*}

\begin{table*}[t]
\centering
\footnotesize
\renewcommand{\arraystretch}{1.18}
\setlength{\tabcolsep}{6pt}

\begin{tabular}{l l l c c c c c c}
\toprule
\textbf{Dataset} & \textbf{Split} & \textbf{Model}
& \textbf{B$>$O} & \textbf{O$>$B}
& \textbf{Net Gain}
& \textbf{McNemar $\chi^2$}
& \textbf{$p$-value}
& \textbf{Significance} \\
\midrule

\multirow{3}{*}{\textsc{RU22Fact}}
& Test & Gemma   & 46 & 246 & +200 & 135.62 & $<10^{-16}$ & \cellcolor{lightgreen}\checkmark \\
& Test & LLaMA   & 44 & 261 & +217 & 152.97 & $<10^{-16}$ & \cellcolor{lightgreen}\checkmark \\
& Test & Mistral & 41 & 253 & +212 & 151.43 & $<10^{-16}$ & \cellcolor{lightgreen}\checkmark \\

\bottomrule
\end{tabular}

\caption{McNemar significance test comparing our SEEK method against the LLM-based baseline.}
\label{tab:appendix_mcnemar_ours_vs_llm}
\end{table*}

\end{document}